\documentclass{article}

\usepackage{microtype}
\usepackage{graphicx}
\usepackage{booktabs} %
\usepackage{amsmath,amsfonts, amsthm, amssymb}
\usepackage{mathtools}
\usepackage{bm} 

\usepackage{algorithm}
\usepackage{algorithmic}

\usepackage{array}
\usepackage{tabularx}
\usepackage{longtable}
\usepackage{ifthen}
\usepackage{makecell}
\usepackage{multirow}
\usepackage{ragged2e}
\usepackage{stfloats}
\usepackage[hypcap=false]{caption}
\usepackage{wrapfig}
\usepackage{ragged2e}

\usepackage[caption=false,font=normalsize,labelfont=sf,textfont=sf]{subfig}

\usepackage{enumitem}
\usepackage{etoolbox}
\usepackage{tcolorbox}
\usepackage{textcomp}
\usepackage{url}
\usepackage{verbatim}
\usepackage{xcolor}
\usepackage{xspace}

\usepackage{hyperref}

\usepackage{iclr2027_conference,times}

\usepackage[capitalise, noabbrev]{cleveref}

\title{Uncovering the Generative Landscape of Diffusion Models: A Spectral-Guided Quality-Diversity Diffusion Framework}

\author{Hannah Janmohamed\\Imperial College London\\ \texttt{hnj21@imperial.ac.uk}
    \And Jo\~ao Correia\\University of Coimbra \\ \texttt{jncor@dei.uc.pt} 
    \And Penousal Machado\\University of Coimbra \\ \texttt{machado@dei.uc.pt}  
    \And Antoine Cully\\Imperial College London \\ \texttt{a.cully@imperial.ac.uk} 
}

\iclrfinalcopy
\begin{document}
\newcommand{\todoinline}[1]{\todo[inline]{#1}}

\newcommand{\codeurl}[0]{\textit{Code hidden for anonymity, will be released upon acceptance.}}

\newcommand{\codepublic}[0]{\url{https://github.com/adaptive-intelligent-robotics/qd-diffusion.git}}

\newcommand{\comparisonBlock}[9]{%
    \def\tempType{#1}%
    \def\tempSeed{#2}%
    \def\tempImgA{#3}%
    \def\tempImgB{#4}%
    \def\tempImgC{#5}%
    \def\tempImgH{#6}%
    \def\tempImgD{#7}%
    \def\tempImgE{#8}%
    \def\tempImgF{#9}%
    \comparisonBlockInternal %
}

\newcommand{\comparisonBlockInternal}[2]{%
    \def\tempImgG{#1}%
    \centering
    \begin{tabular}{@{} p{0.18\textwidth} @{\hspace{0.04\textwidth}} p{0.78\textwidth} @{}}
        \begin{minipage}[t]{0.18\textwidth}
            \vspace{0pt}
            \raggedright \textbf{\scriptsize #2} \\ \vspace{2pt}
            \includegraphics[width=\linewidth]{\tempSeed} \\
            \centering \scriptsize Seed Image
        \end{minipage} & 
        
        \begin{minipage}[t]{0.78\textwidth}
            \vspace{0pt}
            \begin{tabular}{@{} p{0.23\linewidth} p{0.23\linewidth} p{0.23\linewidth} p{0.23\linewidth} @{}}
                \includegraphics[width=\linewidth]{\tempImgA} & 
                \includegraphics[width=\linewidth]{\tempImgB} & 
                \includegraphics[width=\linewidth]{\tempImgC} & 
                \includegraphics[width=\linewidth]{\tempImgH} \\
                \centering\scriptsize\textbf{Ours} & 
                \centering\scriptsize \ifthenelse{\equal{\tempType}{baseline}}{CVT-ME}{Random Seeds} & 
                \centering\scriptsize \ifthenelse{\equal{\tempType}{baseline}}{Threshold}{Diverse Diffusion} & 
                \centering\scriptsize \ifthenelse{\equal{\tempType}{baseline}}{New Baseline}{Particle Guidance} \tabularnewline
            \end{tabular}
            
            \vspace{4pt}
            
            \begin{tabular}{@{} p{0.23\linewidth} p{0.23\linewidth} p{0.23\linewidth} p{0.23\linewidth} @{}}
                \includegraphics[width=\linewidth]{\tempImgD} & 
                \includegraphics[width=\linewidth]{\tempImgE} & 
                \includegraphics[width=\linewidth]{\tempImgF} & 
                \includegraphics[width=\linewidth]{\tempImgG} \\
                \centering\scriptsize \ifthenelse{\equal{\tempType}{baseline}}{Sep-CMA}{ImageBreeder} & 
                \centering\scriptsize \ifthenelse{\equal{\tempType}{baseline}}{CMA-MAE}{Gaussian} & 
                \centering\scriptsize \ifthenelse{\equal{\tempType}{baseline}}{Adam}{Isoline-DD} & 
                \centering\scriptsize \ifthenelse{\equal{\tempType}{baseline}}{QDHF}{Intermed. Gaussian} \tabularnewline
            \end{tabular}
        \end{minipage}
    \end{tabular}
    
    \vspace{0.2cm} \hrule \vspace{0.4cm}
}

\newcommand{\comparisonBlockSmall}[9]{
    \centering
    \begin{tabular}{@{} p{0.18\textwidth} @{\hspace{0.04\textwidth}} p{0.78\textwidth} @{}}
        \begin{minipage}[t]{0.18\textwidth}
            \vspace{0pt}
            \raggedright \textbf{\scriptsize #9} \\ \vspace{2pt}
            \includegraphics[width=\linewidth]{#2} \\
            \centering \scriptsize Seed Image
        \end{minipage} & 
        
        \begin{minipage}[t]{0.78\textwidth}
            \vspace{0pt}
            \begin{tabular}{@{} p{0.28\linewidth} p{0.28\linewidth} p{0.28\linewidth} @{}}
                \includegraphics[width=\linewidth]{#3} & 
                \includegraphics[width=\linewidth]{#4} & 
                \includegraphics[width=\linewidth]{#5} \\
                \centering\scriptsize\textbf{Ours} & 
                \centering\scriptsize CVT-ME & 
                \centering\scriptsize Sep-CMA-ES \tabularnewline
            \end{tabular}
            
            \vspace{4pt}
            
            \begin{tabular}{@{} p{0.28\linewidth} p{0.28\linewidth} p{0.28\linewidth} @{}}
                \includegraphics[width=\linewidth]{#6} & 
                \includegraphics[width=\linewidth]{#7} & 
                \includegraphics[width=\linewidth]{#8} \\
                \centering\scriptsize CMA-MAE & 
                \centering\scriptsize Adam & 
                \centering\scriptsize QDHF \tabularnewline
            \end{tabular}
        \end{minipage}
    \end{tabular}
    
    \vspace{0.2cm} \hrule \vspace{0.4cm}
}

\maketitle

\begin{figure*}[ht]
    \centering
    \begin{minipage}{0.325\textwidth}
        \centering
        \subfloat[\scriptsize random seeds]{%
            \includegraphics[width=0.48\linewidth]{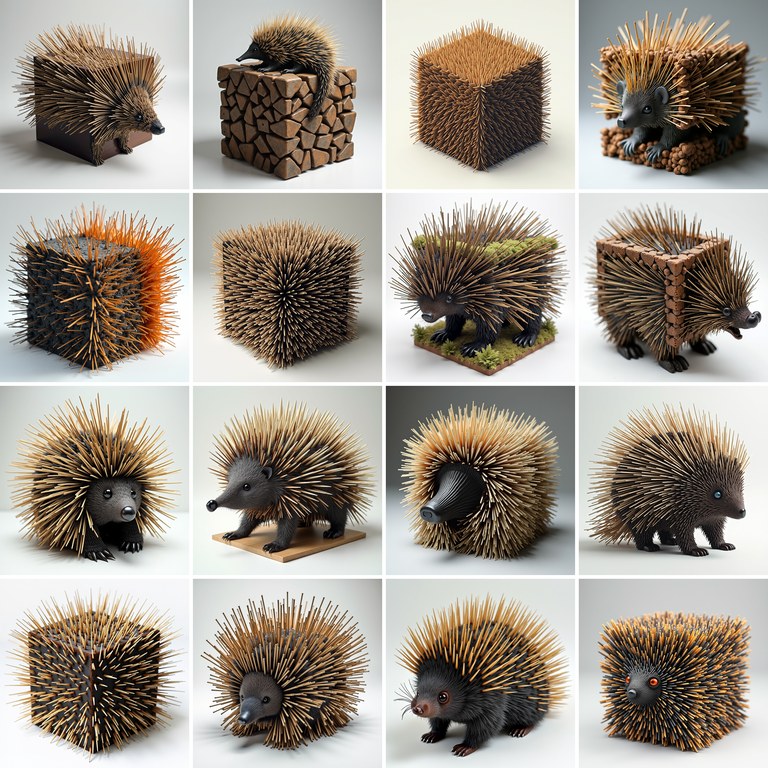}%
        }%
        \hfill
        \subfloat[\scriptsize ours]{%
            \includegraphics[width=0.48\linewidth]{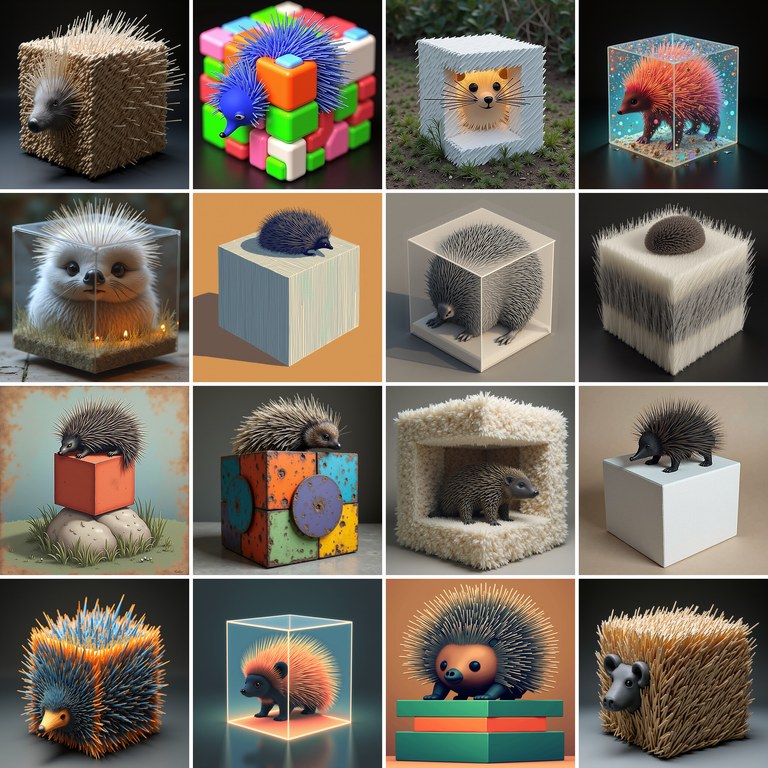}%
        }
    \end{minipage}%
    \hfill 
    \begin{minipage}{0.325\textwidth}
        \centering
        \subfloat[\scriptsize random seeds]{%
            \includegraphics[width=0.48\linewidth]{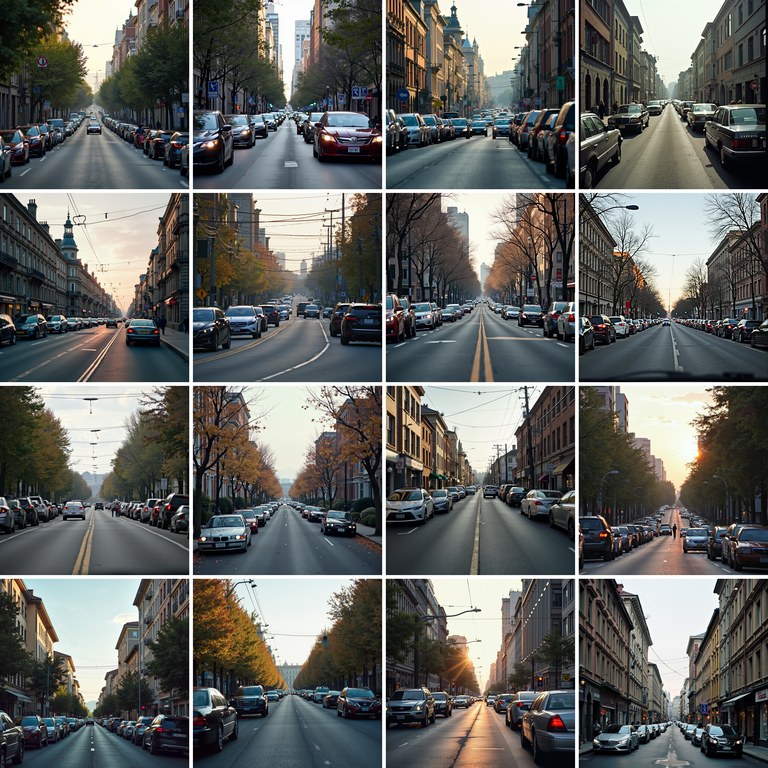}%
        }%
        \hfill
        \subfloat[\scriptsize ours]{%
            \includegraphics[width=0.48\linewidth]{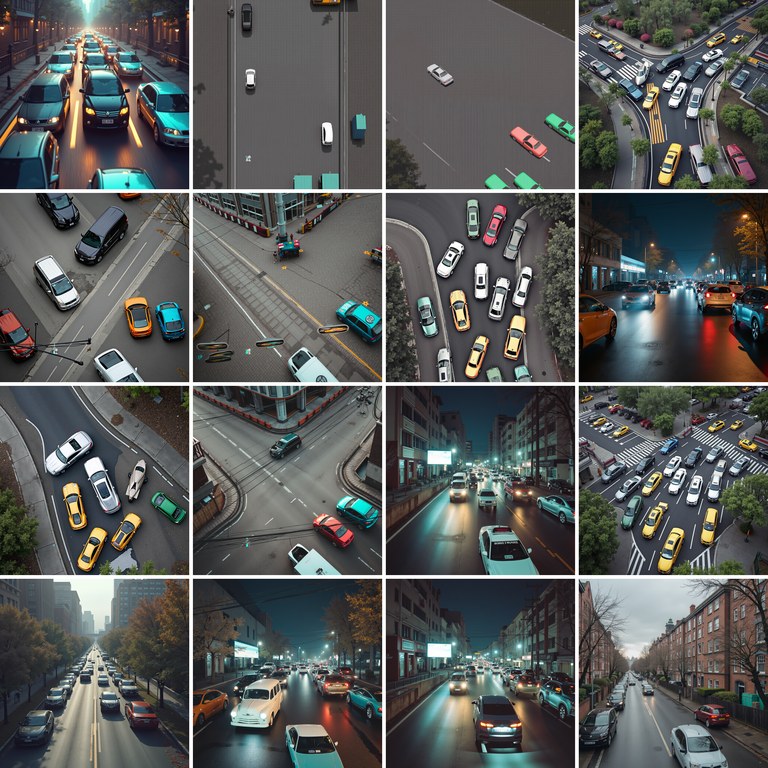}%
        }
    \end{minipage}%
    \hfill 
    \begin{minipage}{0.325\textwidth}
        \centering
        \subfloat[\scriptsize random seeds]{%
            \includegraphics[width=0.48\linewidth]{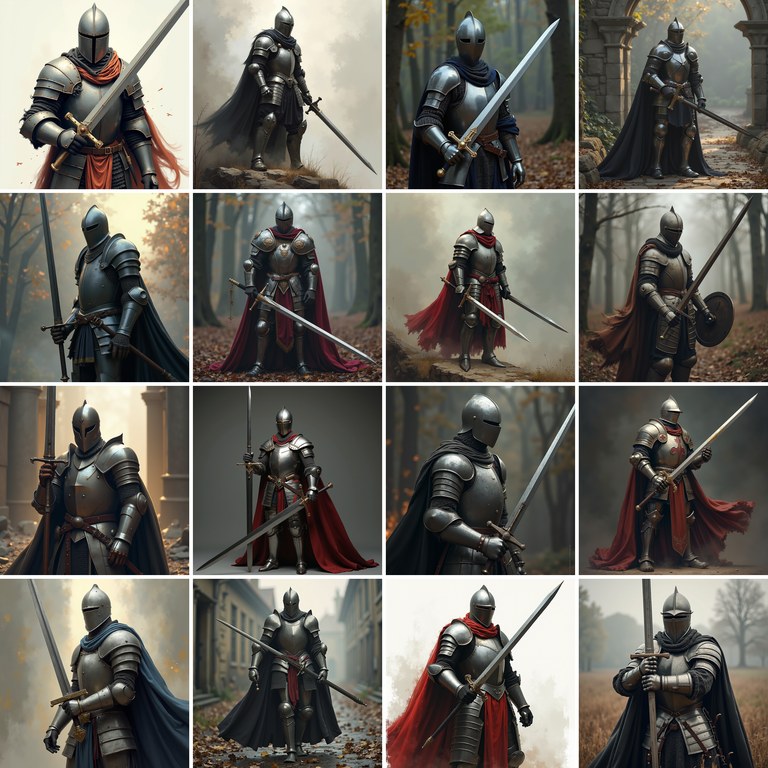}%
        }%
        \hfill
        \subfloat[\scriptsize ours]{%
            \includegraphics[width=0.48\linewidth]{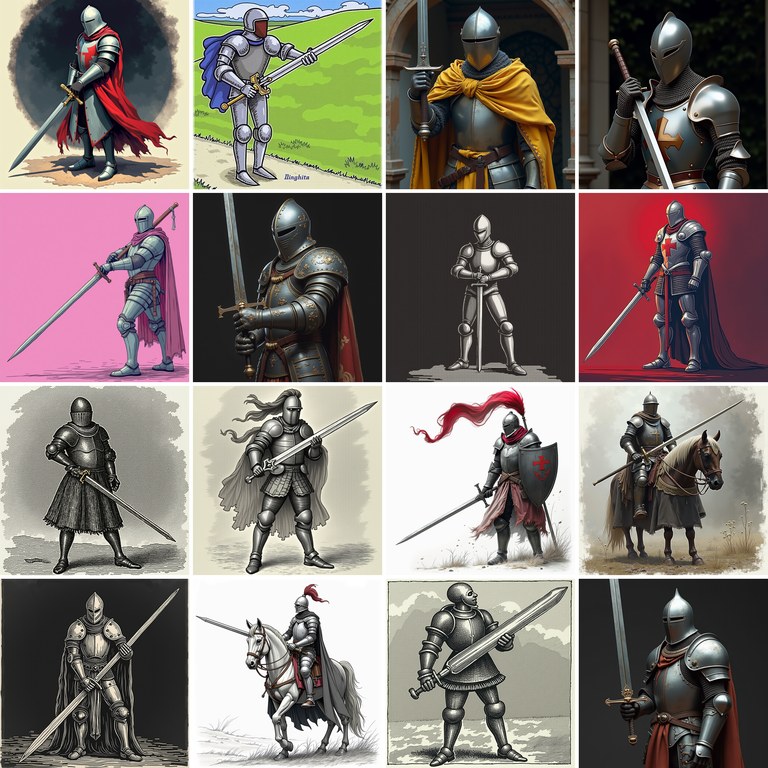}%
        }
    \end{minipage}

    \caption{Flux.1 [dev] \citep{flux2024} outputs under standard sampling versus our Spectral-Guided Quality-Diversity Diffusion. (Left) ``a cube made of porcupine'', (Middle) ``a street with several cars on it'', and (Right) ``a knight holding a long sword''. Random sampling converges to structurally similar modes. Our method exhibits more visual diversity while maintaining prompt alignment.}
    \label{fig:teaser}
\end{figure*}

\begin{abstract}
Denoising Diffusion Probabilistic Models have emerged as powerful generative priors across domains ranging from molecular design to high-resolution image synthesis. 
However, standard diffusion sampling is inherently biased toward high-likelihood regions, often limiting the diversity of outputs.
In many domains, a single, high-likelihood sample is insufficient.
Instead, diverse portfolios of high-performing solutions are required, motivating methods that uncover a model's full expressive range.
Here we show that integrating Quality-Diversity (QD) optimisation with diffusion models enables the discovery of diverse, high-performing outputs.
We first propose \textit{Quality-Diversity Diffusion}, a framework that adapts QD to diffusion models by directly optimising the model's latent noise and conditioning embeddings and leveraging embedding models to quantify diversity.
Second, we introduce a novel \textit{$h$-space Spectral Mutation} operator, which guides search along directions of maximal semantic sensitivity by analysing principal singular vectors of the model's input-to-activation Jacobian.
We evaluate our combined approach, \textit{Spectral-Guided Quality-Diversity Diffusion}, on text-to-image generation.
Against $12$ baselines, our method consistently achieves the lowest FID score, demonstrating an excellent balance of image quality and diversity.
\end{abstract}

\section{Introduction}\label{sec:introduction}
\vspace{-3mm}

Denoising Diffusion Probabilistic Models (DDPMs) \citep{diffusion} have emerged as powerful generative priors, achieving state-of-the-art results across domains from protein design \citep{watson2023novo} to audio synthesis \citep{kong2020diffwave}.
By learning to reverse a gradual noise-adding process, diffusion models approximate the underlying data distribution of a given design domain.

However, standard diffusion sampling procedures are inherently biased towards high-likelihood regions of the learned distribution, often suppressing the diversity inherent in the model's latent space \citep{classifierfree, cads, particle-guidance}.
In many design tasks, a single high-likelihood solution is insufficient: in biological or materials design, candidates that perform well \emph{in silico} may fail under real-world conditions due to modelling errors or the sim-to-real gap.
Instead, identifying a \emph{diverse portfolio} of structurally distinct candidates increases the chance that at least one succeeds in downstream experimental validation \citep{oneshot, moqdcsp, qd4csp}.
Alternatively, in text-to-image generation, diverse outputs support creative exploration by exposing users to a range of possible visual interpretations of a prompt \citep{resnick2005design}.
These examples highlight the need for methods that uncover the full expressive range of a generative prior, rather than just its dominant modes.

In principle, diffusion models generate diverse outputs by resampling the initial noise; in practice, large-scale models remain biased towards the modes of the training distribution, frequently producing artefacts with near-identical global structure across seeds (see \Cref{fig:teaser}).
This calls for a principled mechanism for \emph{exploration} rather than repeated exploitation of high-probability regions -- a role well suited to
Quality-Diversity (QD) optimisation \citep{qdunifying, pugh2016quality}.
Unlike traditional optimisation, which converges to a single global optimum, QD algorithms illuminate the search space returning an archive of high-performing solutions that are behaviourally diverse.

Applying QD optimisation to diffusion-parameterised tasks presents two key challenges.
First, diversity must capture meaningfully semantic or functional differences between artefacts (i.e., images, molecules, audio).
In many complex domains, such notions of diversity are difficult to specify manually and are highly task-dependent.
We address this by leveraging pretrained foundation models, whose embedding spaces act as proxies for distinctiveness between artefacts.
However, their high dimensionality makes them difficult to integrate into existing QD approaches.
Second, the search space induced by a diffusion model is itself high-dimensional, requiring search strategies that can efficiently navigate it while producing meaningful variation in the generated artefacts.

In this work, we introduce \textit{Spectral-Guided Quality-Diversity Diffusion}, a new approach for QD algorithms designed to uncover the hidden diversity of diffusion models.
Our contributions are:

\begin{enumerate}[leftmargin=*, wide=0pt]
    \item \textbf{QD-Diffusion:} We present a Quality-Diversity framework that directly optimises the continuous latent noise and conditional embedding spaces of diffusion models.
    Our approach measures diversity via high-dimensional embeddings from pretrained foundation models.
    By doing so, our framework avoids hand-defined, low-dimensional features required by previous QD approaches, enabling application to domains where explicit axes of diversity are difficult to specify.
    \item \textbf{The $h$-space Spectral Mutation Operator:} We introduce a novel mutation operator tailored to the geometry of diffusion models. Recognising that traditional evolutionary operators in high-dimensional latent spaces are inefficient, our operator computes the principal singular vectors of the model's input-to-activation Jacobian. Perturbing along these directions of maximum semantic sensitivity enables efficient exploration of the latent manifold.
\end{enumerate}

We validate our approach on the challenging domain of text-to-image generation using a representative subset of the Parti Prompts benchmark \citep{partiprompts}.
We test our framework against $5$ distinct baseline algorithms and benchmark our mutation operator against $7$ alternative operators, demonstrating that our method achieves the lowest FID score across all $12$ baselines.
Our code is open-source, available at \codepublic.

\section{Background}
\vspace{-3mm}

\subsection{Diffusion Models}

Denoising Diffusion Probabilistic Models (DDPMs) \citep{diffusion} are a class of generative models that learn to approximate a data distribution $q(\mathbf{x}_0)$ by reversing a gradual noise-adding process. The framework is defined by two Markov chains: a fixed forward process that gradually perturbs data with Gaussian noise until it becomes indistinguishable from isotropic noise, and a parametrised reverse process that learns to denoise it. 

\paragraph{The Reverse Generative Process}
\begin{figure}[ht!]
    \centering
    \includegraphics[width=0.33\linewidth]{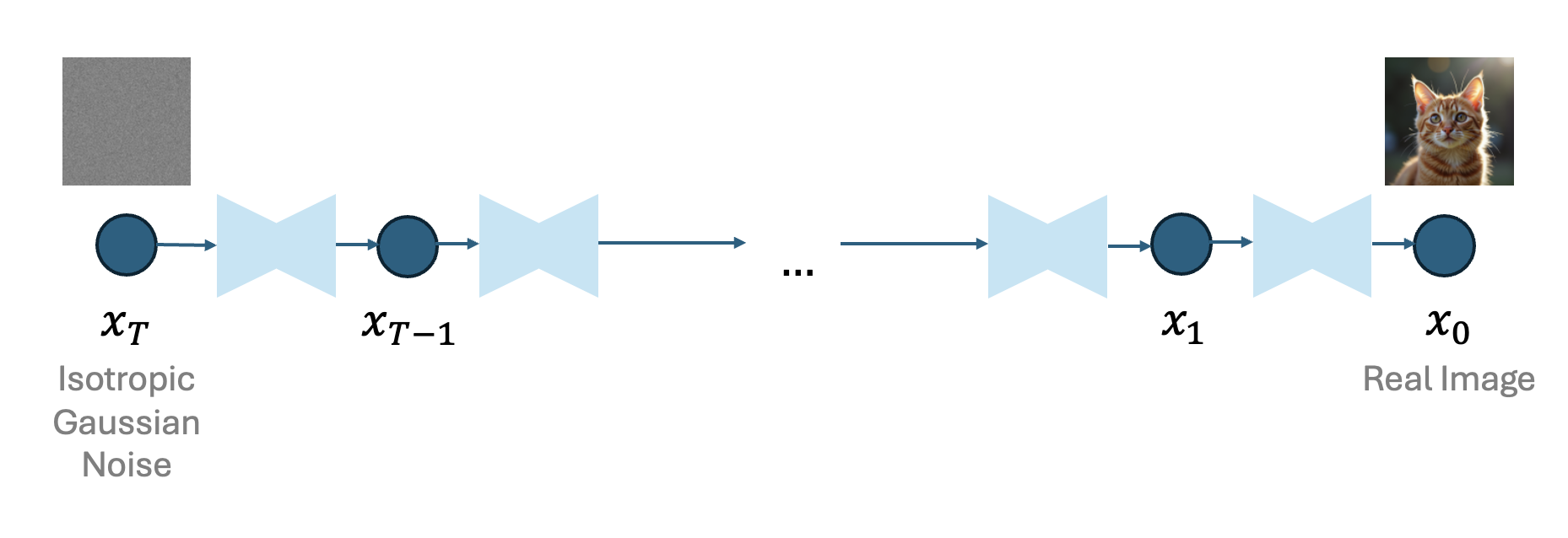}
    \caption{Visualisation of the reverse diffusion process.}
    \label{fig:diffusion_generation}
    \vspace{-2mm}
\end{figure}

Once the model, parametrised by $\theta$, is trained, the core of the generative capability lies in the reverse Markov chain.
To generate an image, we sample pure noise $\mathbf{x}_T \sim \mathcal{N}(\mathbf{0}, \mathbf{I})$ and iteratively denoise it over $T$ timesteps, as visualised in \Cref{fig:diffusion_generation}.

The joint distribution of this reverse process, parametrised by $\theta$, is defined as:
\vspace{-5mm}

\begin{equation}
    p_\theta(\mathbf{x}_{0:T}) = p(\mathbf{x}_T) \prod_{t=1}^T p_\theta(\mathbf{x}_{t-1} | \mathbf{x}_t)\,,
\end{equation}
\vspace{-3mm}

where the transition kernel $p_\theta(\mathbf{x}_{t-1} | \mathbf{x}_t)$ represents a single denoising step. Assuming the noise added at each step of the forward process is sufficiently small, this reverse transition can be approximated by a Gaussian distribution:
\vspace{-2mm}
\begin{equation}
    p_\theta(\mathbf{x}_{t-1} | \mathbf{x}_t) = \mathcal{N}(\mathbf{x}_{t-1}; \boldsymbol{\mu}_\theta(\mathbf{x}_t, t), \boldsymbol{\Sigma}_\theta(\mathbf{x}_t, t))
\end{equation}
At each step, the model estimates the mean $\boldsymbol{\mu}_\theta$ and variance $\boldsymbol{\Sigma}_\theta$ required to transition from the noisy state $\mathbf{x}_t$ to the less noisy state $\mathbf{x}_{t-1}$.
By recursively applying this transition starting from $T$, the model gradually resolves structure, resulting in a sample $\mathbf{x}_0$ from the learned data distribution.
Conditional generation is achieved by injecting context embeddings $c$ into the denoising network via cross-attention or concatenation, guiding outputs toward the desired attributes.

\subsection{Quality-Diversity Algorithms}\label{sec:qd_background}

Quality-Diversity (QD) algorithms aim to illuminate a search space by discovering a collection of solutions that are both high-performing and behaviourally diverse \citep{qdunifying}.
In the QD paradigm, each candidate solution $x$ within the search space $\mathcal{X}$ is evaluated by two functions: a fitness function $f: \mathcal{X} \to \mathbb{R}$, which measures task performance, and a feature function $\phi: \mathcal{X} \to \Phi(\mathcal{X}) \subset \mathbb{R}^d$, which maps the solution to a feature space $\Phi(\mathcal{X})$ where diversity is sought.
The objective is to find a set of solutions $\mathcal{A} = \{x^{(1)}, \dots, x^{(N)}\}$ that maximises $f(x)$ for every unique region or feature in $\Phi(\mathcal{X})$. 
For example, in materials design, the fitness $f$ might measure stability while the feature $\phi$ captures conductivity; QD then seeks a set of stable materials spanning diverse conductivity values \citep{qd4csp, moqdcsp}.

The standard QD framework maintains an archive of solutions organised by their features.
At each iteration, candidate solutions are selected from the archive, subjected to variation operators (mutation or crossover), evaluated, and potentially inserted back into the archive based on an elitist replacement rule \citep{qdunifying}.
A canonical instantiation is MAP-Elites \citep{mapelites}, which discretises the feature space $\Phi(\mathcal{X})$ into a multi-dimensional grid.
A new candidate $x'$ is added to a cell if that cell is empty, or if $f(x')$ exceeds the fitness of the existing occupant $f(x)$.
This mechanism ensures the archive monotonically improves in quality while maintaining diversity across the grid.
We provide the pseudocode for MAP-Elites in Appendix \ref{app:me_pseudo}.

While effective in low-dimensional settings, the grid-based discretisation of MAP-Elites becomes inefficient as the dimensionality of $\Phi(\mathcal{X})$ increases.
Dominated Novelty Search (DNS) \citep{dns} overcomes this by maintaining an unstructured archive.
At each generation, both offspring and archived solutions are ranked using a scoring metric that balances fitness and novelty.
Specifically, a solution's score is calculated as the average distance in feature space to its $k$-nearest neighbours that possess higher fitness.
This logic ensures that solutions are preserved if they are either distinct (isolated in feature space) or high-performing (few neighbours have better fitness).
Because DNS relies on distance metrics rather than grid discretisation, it scales effectively to high-dimensional feature spaces.
The pseudocode for DNS is provided in Appendix \ref{app:dns_pseudo}.

\section{Methodology}
\vspace{-3mm}

\subsection{Framework}\label{sec:method_framework}
\vspace{-2mm}
\paragraph{Overview}
In this section, we detail the \textit{Quality-Diversity Diffusion framework} which integrates diffusion models with evolutionary search to discover diverse, high-performing designs.
We assume the existence of a design task defined by:
\begin{enumerate}[leftmargin=*]
    \item A pre-trained diffusion model $p_\theta(x_{0:T})$ capable of generating data samples.
    \item A fitness function $f: \mathcal{X} \to \mathbb{R}$ that evaluates the quality or task performance of a generated sample.
    \item An embedding model $\phi: \mathcal{X} \to \mathbb{R}^d$ that maps samples to a feature space representing semantic or behavioural characteristics. 
\end{enumerate}

\begin{wrapfigure}{l}{0.5\textwidth}
    \centering
    \vspace{-15pt} %
    \includegraphics[width=0.48\textwidth]{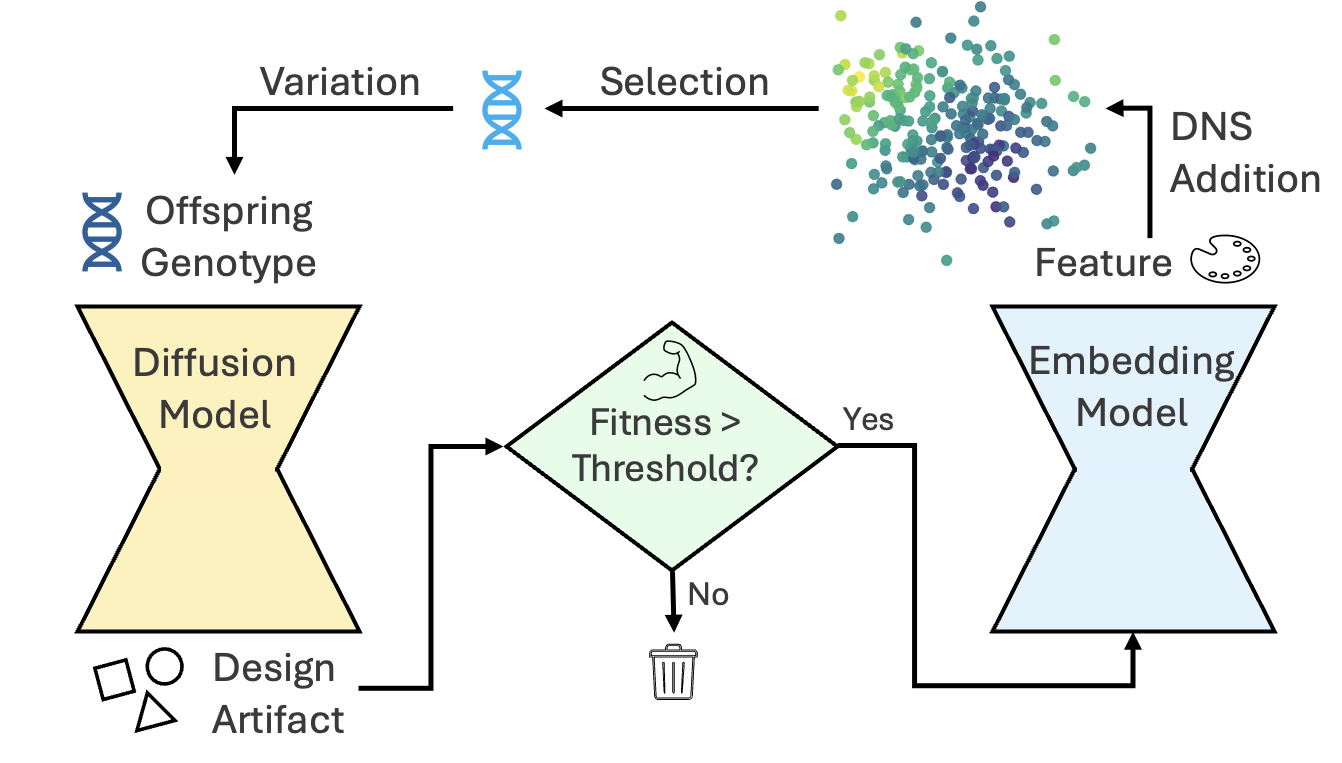}
    \caption{Quality-Diversity Diffusion Framework.}
    \label{fig:framework}
    \vspace{-10pt}
\end{wrapfigure}

{\RaggedRight Our framework follows a standard Quality-Diversity loop (see \Cref{fig:framework}). Genotypes (e.g., latent noise $x_T$ and conditioning embeddings $c$) are selected and varied to produce offspring. In this work, we propose the $h$-space Spectral Mutation Operator although any variation operator can be used. The diffusion model maps these to new artifacts, which are evaluated for fitness, embedded in a high-dimensional feature space and added to the unstructured archive using Dominated Novelty Search with fitness thresholding. \par}

\paragraph{Genotypes and Mutations}

We define the genotype as the tuple $(x_T, c)$, where $x_T \sim \mathcal{N}(0, I)$ is the initial Gaussian noise vector and $c$ is the conditioning embedding vector.
Restricting our search space to these variables ensures the framework is domain-agnostic and applicable to any diffusion-based task (e.g., images, proteins, audio) without requiring text-specific linguistic priors.

\paragraph{DNS With Fitness Thresholding}

Standard QD algorithms such as MAP-Elites often rely on low-dimensional, hand-crafted features \citep{book_chapter, qdunifying, mapelites} which are often unavailable in complex domains.
In this work, we rely on the assumption that an increasing number of design domains have access to rich representations via pre-trained embedding models (e.g., CLIP for images \citep{clipscore}, ESM for proteins \citep{lin2023evolutionary}), which act as powerful proxies for semantic distinctiveness.
However, the high dimensionality of these vectors (often hundreds of dimensions) renders grid-based algorithms infeasible.

To address this, we employ Dominated Novelty Search (DNS) as the underlying archive mechanism.
As explained in \Cref{sec:qd_background} DNS maintains an unstructured archive, calculating a score for each solution based on the average distance to its $k$-nearest fitter neighbours in feature space. 
While DNS has been shown to scale well to high-dimensions, we identified a critical limitation when applying it to generative tasks. 
In particular, since the DNS scoring mechanism balances the quality and novelty of a solution, we found that extremely novel solutions were valued highly regardless of their validity.
In generative design, mutations in the latent space can easily traverse boundaries into invalid regions of the manifold (e.g., generating a live ``turkey'' instead of a ``roast turkey''); standard DNS erroneously preserves these failures due to their high novelty scores.
Hence, degenerate solutions are frequently retained and then propagated through the archive solely because they are novel.
To mitigate this, we introduce a fitness-thresholding mechanism.
A candidate solution $x'$ is only considered for addition to the archive if its fitness exceeds a minimum threshold, i.e., $f(x') > f_{\text{thresh}}$.
This restricts search to valid regions of the generative manifold.
An ablation study on this mechanism is provided in Appendix \Cref{app:threshold_ablation}.

\subsection{\texorpdfstring{\textit{$h$}}{h}-space Spectral Mutation Operator}\label{sec:method_variation}

Given the framework outlined in \Cref{sec:method_framework}, the primary challenge shifts to efficient exploration within the vast latent space of diffusion models.
To address this, we introduce the \textbf{$h$-space Spectral Mutation Operator}.

\paragraph{Spectral Sensitivity Analysis}

Our operator aims to traverse the latent space along meaningful axes, which requires extracting the latent structure of the input space $\mathcal{X}_t$.
Directly analysing $\mathcal{X}_t$ is difficult, however, since diffusion training does not enforce a globally organised semantic structure, and the geometry of $\mathcal{X}_t$ varies with timestep $t$ \citep{riemannian_geometry_latent_space}.

We instead turn to the model's internal bottleneck representations, or \textit{hidden space} $\mathcal{H}$ ($h$-space), which are known to preserve local linear structure corresponding to semantic features \citep{kwon2022diffusion}.
Let $g : \mathcal{X}_t \rightarrow \mathcal{H}$ denote the mapping from the latent at timestep $t$ to this hidden space.
Since $g$ is non-linear, we characterise its local geometry via \textit{spectral analysis} of its Jacobian, using differential geometry to approximate its behaviour between tangent spaces $\mathcal{T}_\mathcal{X}$ and $\mathcal{T}_\mathcal{H}$.
Following \citet{riemannian_geometry_latent_space}, the Jacobian $J_g(\mathbf{x}) = \nabla_{\mathbf{x}} g(\mathbf{x})$ defines a linear map $\mathcal{T}_\mathcal{X} \rightarrow \mathcal{T}_\mathcal{H}$, letting us identify the perturbation $\mathbf{v} \in \mathcal{T}_\mathcal{X}$ that induces maximal variance in $\mathcal{T}_\mathcal{H}$:
\vspace{-1mm}
\begin{equation}
    \mathbf{v}^* = \mathop{\mathrm{arg\,max}}_{\mathbf{v} : \|\mathbf{v}\|_2=1} \| J_g(\mathbf{x}) \mathbf{v} \|^2
\end{equation}
$\mathbf{v}^*$ is the first right singular vector of $J_g(\mathbf{x})$ (i.e., the first column of $V$ in $J_g(\mathbf{x}) = U\Sigma V^T$, satisfying $J_g(\mathbf{x})\mathbf{v}^* = \sigma_1 \mathbf{u}_1$): moving along it produces the largest possible change in the model's internal features for a fixed step size.

Based on this analysis, we define our mutation operator as a perturbation of the latent state $\mathbf{x}_{t}$ along this principal semantic direction:
\vspace{-3mm}
\begin{equation}
    \tilde{\mathbf{x}}_{t} = \mathbf{x}_{t} + \alpha \mathbf{v}^*
\end{equation}
where $\alpha \sim \text{Unif}[\alpha_{\min} \alpha_{\max}]$, is the step size.
By aligning perturbations with these principal axes, we force the search along the most semantically active regions of the latent manifold, avoiding the inefficiencies of unguided evolutionary variation operators.
We note that this approach is architecture-agnostic: $g(\mathbf{x}; t, c)$ is a generic input-to-feature mapping, corresponding to an intermediate convolutional block in U-Net models and a transformer layer in DiT models, so the analysis applies regardless of backbone.

Furthermore, we extend the analysis of \citet{riemannian_geometry_latent_space} beyond the latent noise $\mathbf{x}_{t}$, by applying the same spectral analysis to the conditioning vectors $c$, identifying directions of maximal sensitivity in the conditional embedding space.
This allows our framework to treat conditioning vectors as mutable genotypes alongside $\mathbf{x}_T$, enabling joint optimisation of the seed and the conditional embeddings.

\begin{figure}[ht]
    \centering
    \includegraphics[width=0.4\linewidth]{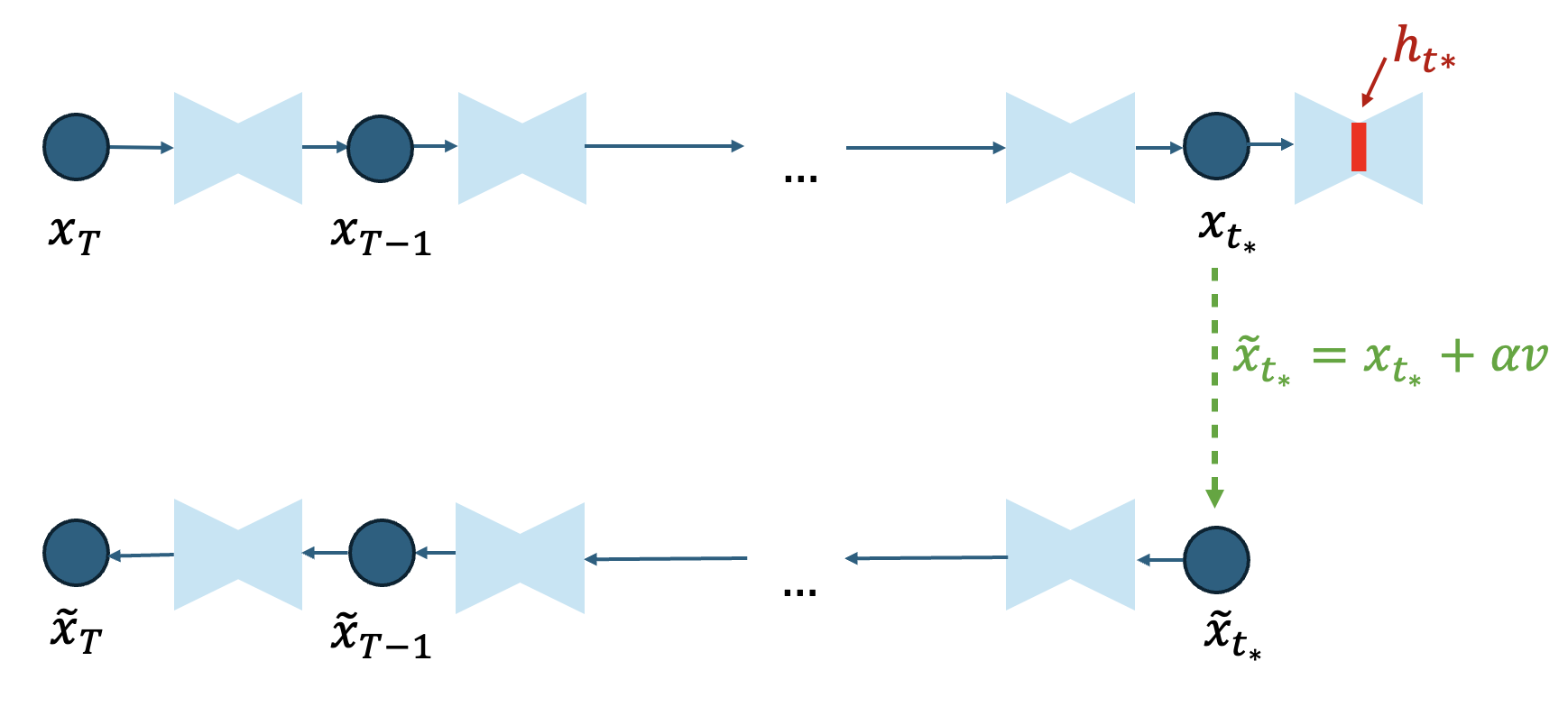}
    \caption{The $h$-space Spectral Mutation Operator. Initial noise $\mathbf{x}_T$ is propagated to a timestep $t_*$ where singular vectors $\mathbf{v}$ of the Jacobian mapping the latent $\mathbf{x}_t$ to the model's internal activations $h$ are computed. The latent is perturbed along this direction ($\mathbf{\tilde{x}}_t = \mathbf{x}_t + \alpha \mathbf{v}$) and the result $\mathbf{\tilde{x}}_t$ is deterministically inverted to the timestep to form the offspring genotype $\mathbf{\tilde{x}}_T$.}
    \label{fig:mutation}
\end{figure}

\paragraph{Deterministic Inversion}
Evolutionary search on generative models is typically performed directly on the initial noise vector space $\mathcal{X}_T$, i.e., at $t=T$ \citep{imagebreeder}.
However, literature in diffusion editing suggests that manipulating latent variables at this initial stage ($t=T$) rarely yields meaningful semantic shifts.
Instead, semantic editing is most effective at intermediate timesteps (typically $t_* \in [0.5T, 0.7T]$), where internal representations are most distinct \citep{riemannian_geometry_latent_space}.
However, we cannot directly store intermediate latents in the archive, since genotypes from different diffusion timesteps would be incompatible for subsequent parent selection.
We resolve this by mutating at an intermediate timestep $t_*$ and using deterministic noise schedulers to invert the modified latent back to the initial noise space $\mathbf{x}_T$, keeping the genotype consistent for archiving.
The overall process, visualised in \Cref{fig:mutation}, proceeds as follows:

\begin{enumerate}[leftmargin=*, wide=0pt]
    \item \textbf{Forward Simulation:} The parent genotype $\mathbf{x}_T$ is propagated to an intermediate timestep $t_*$ (randomly sampled or a hyperparameter) using a deterministic discrete scheduler such as DDIM or Euler Discrete \citep{diffusion, karras2022elucidating}. In these deterministic settings, the variance term $\boldsymbol{\Sigma}_\theta$ (which typically controls stochastic noise injection) is effectively zero, ensuring that the trajectory is bijective. The intermediate latent is thus obtained by iteratively applying the scheduler's deterministic update rule, using only the predicted mean $\boldsymbol{\mu}_\theta$:
    \vspace{-1mm}
    \begin{equation}
        \mathbf{x}_{s-1} = \boldsymbol{\mu}_\theta(\mathbf{x}_s, s) \quad \text{for } s = T, \dots, t_* + 1
    \end{equation}
    \vspace{-1mm}
    \item \textbf{Perturbation:}
    Since exact computation of $J_g(\mathbf{x})$ is intractable, we follow \citet{riemannian_geometry_latent_space} and employ a \textit{Power Iteration} method leveraging finite differences and automatic differentiation to approximate the top-$k$ singular vectors $\{\mathbf{v}_1, \dots, \mathbf{v}_k\}$ of the mapping $g$'s Jacobian. We compute these sensitivity directions for a single parent and create a batch of $b$ new solutions by sampling different scalings $\alpha$ and vectors $\{\mathbf{v}_1, \dots, \mathbf{v}_k\}$:
    \vspace{-3mm}
    \begin{equation} \label{eqn:mutation}
        \tilde{\mathbf{x}}_{t_*} = \mathbf{x}_{t_*} + \alpha \mathbf{v}
    \end{equation}
    where $\alpha$ is a scalar step size sampled uniformly from $[\alpha_{\min}, \alpha_{\max}]$ and $\mathbf{v}$ is randomly chosen from $\{\mathbf{v}_1, \dots, \mathbf{v}_k\}$, reusing the expensive gradient computation across the whole offspring batch.   
    \item \textbf{Deterministic Inversion:} %
    Since deterministic schedulers define a bijective mapping between noise levels, we can compute the inversion by iteratively applying the inverse update rule:
    \vspace{-3mm}

    \begin{equation}
        \tilde{\mathbf{x}}_{s+1} = \boldsymbol{\mu}_\theta^{-1}(\tilde{\mathbf{x}}_s, s) \quad \text{for } s = t_*, \dots, T - 1
    \end{equation}
    translating semantic changes at $t_*$ back into the noise space $\mathbf{x}_T$. This step is unnecessary for distilled, one-step models, where perturbations apply directly to the input noise.
\end{enumerate}

\vspace{-3mm}
\section{Experimental Setup}\label{sec:setup}
\vspace{-3mm}

\subsection{Task}

\vspace{-2mm}
\paragraph{Domain} We assess the efficacy of our method in the domain of text-to-image generation.
We conduct our primary evaluation on SDXL-Turbo \citep{sdxlturbo}, selected for its efficiency.
To ensure our findings hold across diverse semantic contexts, we utilise the Parti Prompts benchmark \citep{partiprompts}.
For SDXL-Turbo, we sample three prompts from each of the provided categories (e.g., abstract, complex), resulting in a diverse set of 36 evaluation tasks.
To demonstrate the generality of our framework, we conduct additional experiments using the FLUX.1 [dev] model \citep{flux2024} which utilises a Diffusion Transformer (DiT) architecture and relies on a multi-step generation process (typically 30-50 steps), in contrast to the U-Net-based, single-step generation of SDXL-Turbo.
Due to the substantial computational cost associated with the Flux model, we restricted this evaluation to a subset of $12$ prompts.
Further implementation details are provided in Appendix \Cref{app:implementations}.

\vspace{-3mm}

\paragraph{Fitness Definition}
The fitness is the fidelity of the images to the prompt, quantified via the CLIP score \citep{clipscore}, a standard metric for image-text alignment.

\vspace{-3mm}

\paragraph{Feature Definition}
We use DreamSim \citep{dreamsim} for our features. The model generates 768-dimensional embeddings trained to approximate human similarity judgements, thereby serving as a proxy for semantic distinctiveness.

\vspace{-3mm}

\paragraph{Genotype}
As noted in \Cref{sec:method_variation}, the spectral variation operator is general and can operate on any genotype. For each model, we define the genotype to encompass all available input components, splitting the batch size equally among them. We apply the same genotype definitions to all applicable baselines. An ablation study isolating each genotype component is provided in \Cref{app:genotype_ablation}.

\subsection{Baselines}\label{sec:baselines}
\vspace{-2mm}

An overview of each baseline is provided below. A comprehensive related works and implementation details of baselines are provided in Appendices \Cref{app:related} and \Cref{app:baselines} respectively.

\vspace{-2mm}
\paragraph{Framework Baselines}
We compare the QD-Diffusion framework design to the following baselines:

\begin{itemize}[leftmargin=*]
    \item \textbf{CVT-ME:} A variant of MAP-Elites designed to scale to high-dimensional spaces by using Centroidal Voronoi Tessellations to discretise the feature space into a fixed number of cells \citep{cvt}.
    \item \textbf{CMA-ES:} A robust evolutionary optimisation strategy designed for high-dimensional spaces \citep{cma-es}. Unlike QD algorithms, CMA-ES does not use a feature space and aims solely to maximise fitness.
    \item \textbf{CMA-MAE:} A state-of-the-art Quality-Diversity algorithm that combines CMA-ES with the MAP-Elites archiving mechanism \citep{cma-mae}. While traditionally reliant on hand-defined feature spaces, we adapt this baseline to use a CVT-ME grid over DreamSim embeddings to ensure a fair comparison within our automated, high-dimensional feature setting.
    \item \textbf{Adam:} Gradient-based optimisation of the latent variable over the fitness function \citep{adam}.
    \item \textbf{QDHF:} Quality-Diversity via Human Feedback \citep{qdhf}. This method uses CLIP embeddings as features and a learned projection layer to map high-dimensional embeddings to a low-dimensional latent space suitable for standard MAP-Elites. The learned layer is trained using a contrastive loss for similarity, with DreamSim serving as a proxy for human judgements.
\end{itemize}

The CVT-ME, CMA-MAE and QDHF baselines are selected to determine whether our framework is the optimal QD algorithm for high-dimensional feature embeddings.
Crucially, CVT-ME and QDHF employ our proposed spectral mutation operator, ensuring that comparisons isolate the efficacy of the archive mechanism rather than the variation strategy.
CMA-ES and Adam represent pure fitness-maximising algorithms, while serving to test whether an explicit diversity mechanism is necessary or if powerful optimisers can implicitly discover diversity.
Adam and QDHF are not evaluated on Flux due to computational constraints (see Appendix \Cref{app:baselines}).

\paragraph{Variation Operator Baselines}
To isolate the efficacy of the $h$-space Spectral Mutation operator from the framework, we benchmark it against the following variation operators:
\vspace{-2mm}

\begin{itemize}[leftmargin=*]
    \item \textbf{Random seeds:} A baseline wherein each iteration samples a batch of solutions via new random seeds, representing the standard usage pattern of diffusion models.
    \item \textbf{Diverse-Diffusion:} A recent approach designed to enhance image variety by performing a search for latent vectors with maximum pairwise distance \citep{diversediffusion}.
    \item \textbf{Particle Guidance:} generates a batch of samples jointly rather than i.i.d., adding a guidance term that repels particles from one another to enforce within-batch diversity \citep{particle-guidance}.
    \item \textbf{ImageBreeder:} An evolutionary operator originally designed for image generation that performs latent-space crossover using a variety of strategies, including weighted blending and slicing along channels or spatial dimensions \citep{imagebreeder}. This is the only operator that is only applied to the latent noise vectors, as the operator was developed specifically for latent-space crossover.
    \item \textbf{Gaussian Noise:} Adds isotropic perturbations to the genotype, defined as $\mathbf{x}' = \mathbf{x} + \sigma \boldsymbol{\epsilon},$ where $\boldsymbol{\epsilon} \sim \mathcal{N}(\mathbf{0}, \mathbf{I})$ and $\sigma$ is the mutation strength hyperparameter. 
    \item \textbf{Isoline-DD:} Generates offspring by perturbing a parent $\mathbf{x}_1$ along the direction relative to a second parent $\mathbf{x}_2$, defined as $\mathbf{x}' = \sigma_1 \mathbf{x}_1 + \sigma_2 \mathcal{N}(0, 1) (\mathbf{x}_1 - \mathbf{x}_2)$ where $\sigma_1$ and $\sigma_2$ are hyperparameters \citep{isoline}.
    \item \textbf{Intermediate-Layer Gaussian noise:} Instead of performing SVD-based updates at the intermediate timestep, we apply Gaussian noise to the intermediate vector $\mathbf{x}_t$ and then invert the process using the deterministic scheduler. 
\end{itemize}

The Random Seeds baseline provides a naive reference point; by contrast, Diverse Diffusion and Particle Guidance are sampling-based methods explicitly designed to promote diversity, and serve as non-evolutionary benchmarks.
Since Particle Guidance operate through classifier-free guidance, it is inapplicable to one-step models and are therefore only evaluated on Flux.
ImageBreeder allows for comparison to diffusion-specific evolutionary operators, while Gaussian Noise and IsoLine represent standard evolutionary baselines.
The Intermediate-Layer Gaussian baseline is critical for demonstrating that meaningful diversity results from spectral mutation specifically, rather than performing mutation at intermediate layers.

\vspace{-2mm}
\paragraph{Hyperparameters}

For SDXL-Turbo, a computational budget of 2000 evaluations is allocated to all algorithms; for QD algorithms, this is processed as batches of 16 over 125 iterations.
Flux experiments used a reduced computational budget of 400 total evaluations, processed as $100$ iterations with a batch size of $4$.
Statistical significance is determined using a Wilcoxon signed-rank test with Holm-Bonferroni correction \citep{wilcoxon1992individual, holm_bonf}.
Further hyperparameter details and an ablation study on these choices are provided in Appendices \Cref{app:global_hyperparams} and \Cref{app:ablations} respectively.

\vspace{-2mm}
\subsection{Evaluation Metrics}
\vspace{-2mm}

Our primary metric is the Fréchet Inception Distance (FID) which is a widely-used metric that measures both the quality and diversity of solutions, compared to a set of real images \citep{fid}. This metric is not reported for algorithms that fail to discover a significant number of solutions. We provide further implementation details in Appendix \Cref{app:fid_implementation}. Since the FID metric aggregates the Fitness and Diversity of solutions into one, we provide further insight via the following metrics:
\vspace{-2mm}

\begin{itemize}[leftmargin=*]
    \item \textbf{Total Fitness:} The total fitness scores of all solutions in the final collection, reflecting both the quality of individual solutions and the algorithm’s ability to populate the archive with solutions exceeding the fitness threshold.
    \item \textbf{Diversity (Mean Pairwise Distance):} The average Euclidean distance between all pairs of embeddings in the final collection, quantifying the semantic spread of the discovered portfolio.
     \item \textbf{Number of Solutions:} The total count of distinct solutions maintained in the final archive.
\end{itemize}

\vspace{-3mm}
\section{Results}
\vspace{-3mm}

\subsection{Framework Results}\label{sec:framework_results}
\vspace{-3mm}

We first evaluate our framework using the FID score, as reported in \Cref{tab:fid_framework}.
Our Spectral-Guided approach achieves the lowest FID scores across both SDXL-Turbo ($81.90$) and Flux ($137.58$) models.
As FID measures the distance between the distribution of generated images and real-world data, these results confirm that our method produces high-quality, realistic outputs.
Notably, baseline methods such as CVT-ME and CMA-MAE failed to discover a sufficient number of solutions across all runs for a valid FID metric to be calculated.

\begin{table}[htbp]
\centering
\tiny %
\setlength{\tabcolsep}{2pt}
\renewcommand{\arraystretch}{1.2}

\begin{minipage}{0.4\textwidth}
\centering
\caption{FID Score.}
\label{tab:fid_framework}
\vspace{-10pt}
\begin{tabularx}{\linewidth}{|X|c|c|}
\hline
\textbf{Algorithm} & \textbf{SDXL-Turbo }& \textbf{Flux}\\
\hline
Spectral-Guided (Ours) & \textbf{81.90}  & \textbf{137.58 }\\
\hline
CVT MAP-Elites & n/a & n/a \\
\hline
Sep-CMA-ES & 145.13 & 196.67\\
\hline
CMA-MAE & n/a & n/a \\
\hline
Adam & 163.70 & - \\
\hline
QDHF & 122.20 & - \\
\hline
\end{tabularx}
\end{minipage}
\hfill
\begin{minipage}{0.4\textwidth}
\centering
\caption{Number of Solutions.}
\label{tab:framework_counts}
\vspace{-10pt}
\begin{tabularx}{\linewidth}{|X|c|c|}
\hline
\textbf{Algorithm} & \textbf{SDXL-Turbo} & \textbf{Flux} \\
\hline
Spectral-Guided (Ours) & \textbf{62.7 $\pm$ 8.0} & \textbf{58.0 $\pm$ 17}\\
\hline
CVT MAP-Elites &  6.9 $\pm$ 3.1  & 5.25 $\pm$ 3.4 \\
\hline
Sep-CMA-ES & \textbf{61.4 $\pm$ 10.7}  & \textbf{54.0 $\pm$ 23} \\
\hline 
CMA-MAE & 8.28 $\pm$ 2.3 & 5.92 $\pm$ 1.8\\
\hline
Adam & \textbf{64.0 $\pm$ 0.0}  & - \\
\hline
QDHF & 48.8 $\pm$ 7.8  & - \\
\hline
\end{tabularx}
\end{minipage}
\end{table}
\vspace{-3mm}

\begin{figure}[t]
  \centering
  \subfloat[SDXL-Turbo.\label{fig:framework_pareto_turbo}]{%
    \includegraphics[width=0.4\textwidth]{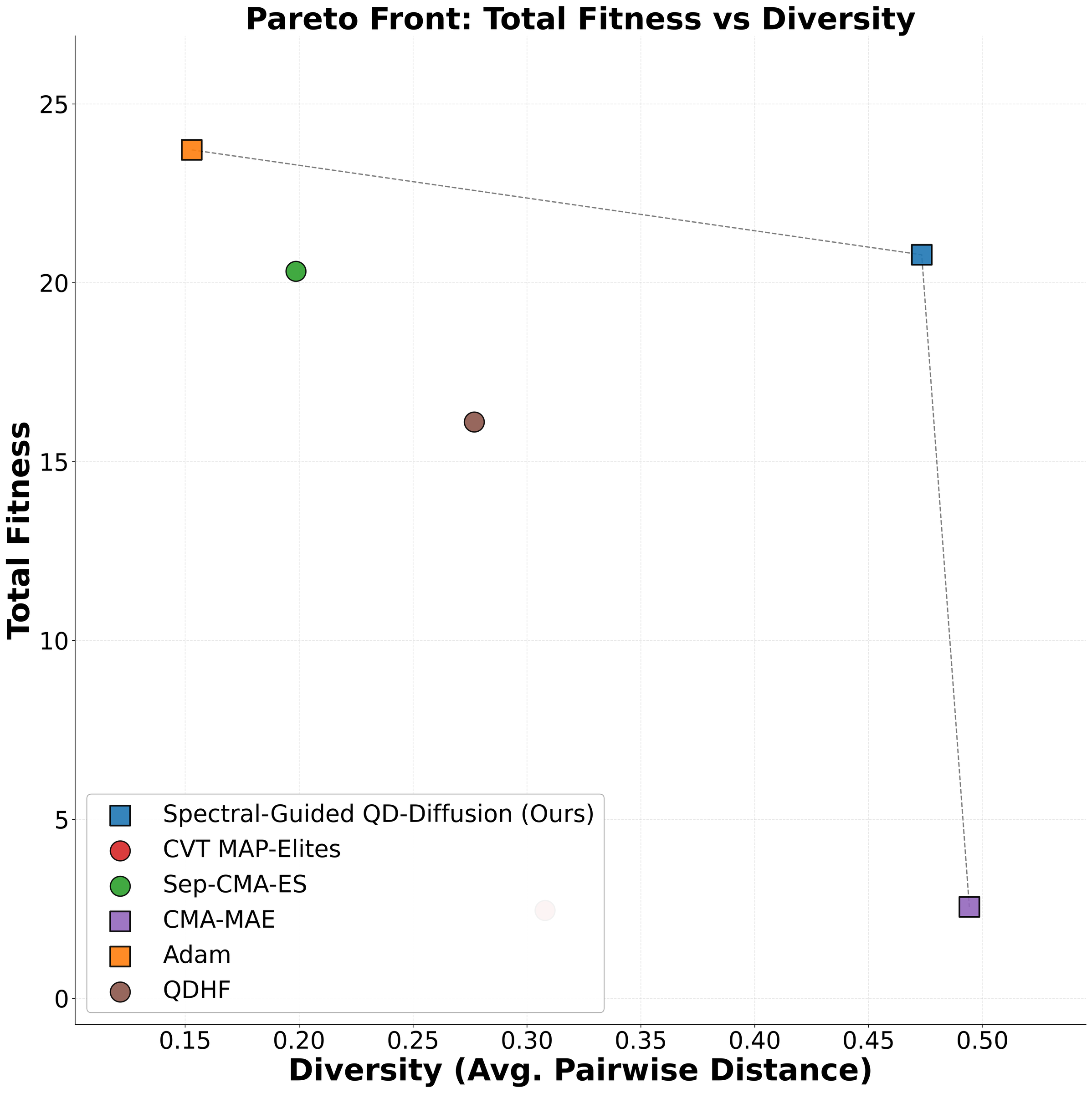}}
  \hspace{2em}
  \subfloat[Flux.\label{fig:framework_pareto_flux}]{%
    \includegraphics[width=0.4\textwidth]{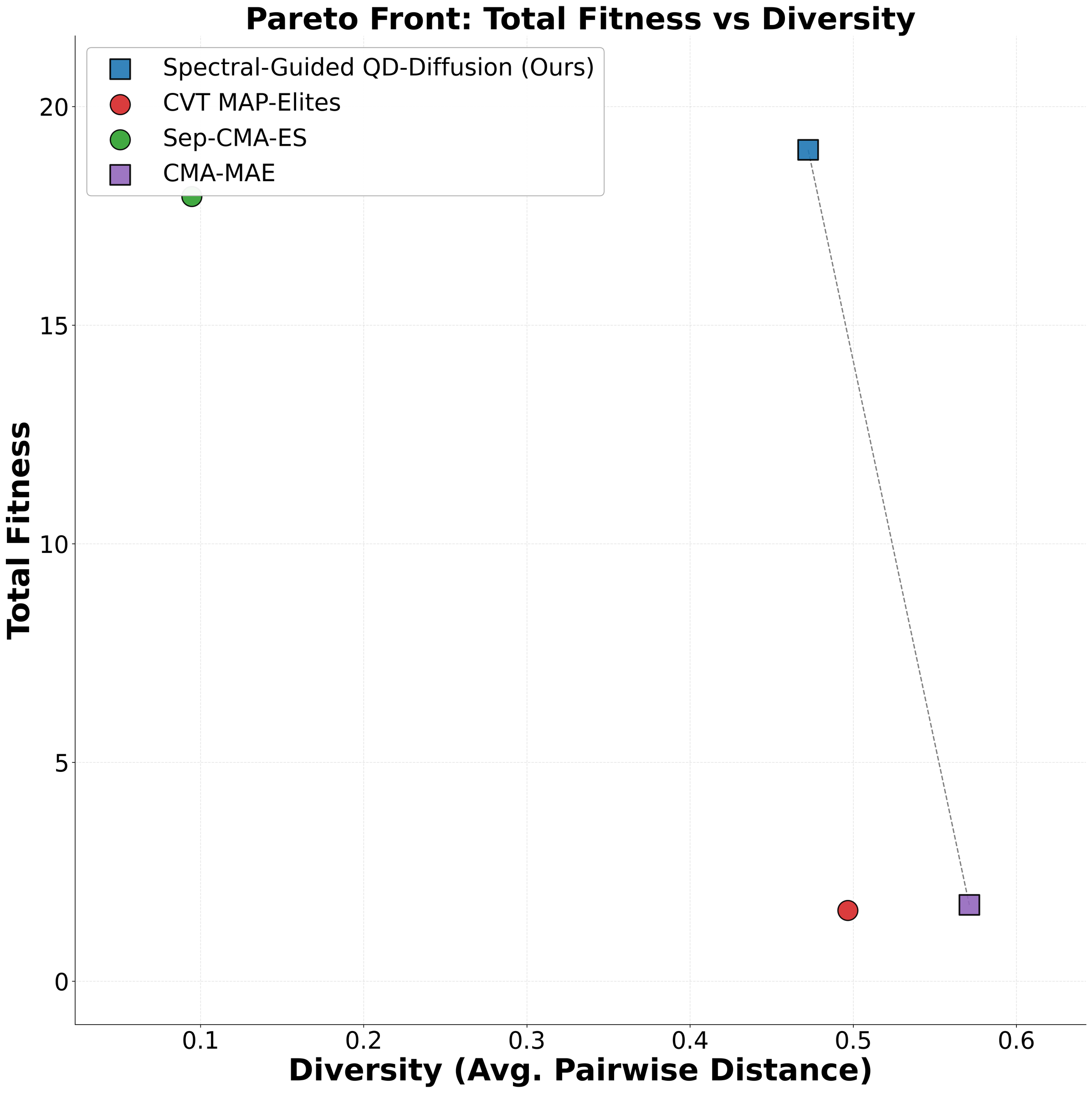}}
  \caption{Quality vs Diversity.}
  \label{fig:framework_pareto}
\end{figure}

To disentangle the contributions of diversity and quality, we examine the Total Fitness and Diversity metrics visualised via the Pareto Front in \Cref{fig:framework_pareto}.
Compared to CVT-ME, our method exhibits clear Pareto domination on SDXL-Turbo, which attains lower Total Fitness ($2.45$ vs $20.8$, $p < 2 \times 10^{-10}$) and a lower Diversity ($0.31$ vs $0.47$, $p < 8 \times 10^{-7}$) than our method.
On Flux, CVT-ME achieves a marginally higher diversity ($0.497$ vs $0.472$) but the difference is not statistically significant ($p=1.0$) and the total fitness is still substantially lower $19.5$ vs $1.97$, $p < 3.5 \times 10^{-2}$.
This performance is attributable to its inability to densely populate the solution space, producing only $6.9$ solutions per prompt on average and highlighting the limitations of grid-based QD methods in high-dimensional feature spaces.

Similarly, CMA-MAE also suffers from the grid-based approach, achieving a significantly lower Total Fitness on both models ($p<4\times10^{-3}$).
While it records higher diversity than our method, the difference is not statistically significant ($p=0.47$ for SDXL-Turbo and $p=0.389$ for Flux) and we hypothesise that this is largely a byproduct of poor image quality.
As shown in the qualitative results in Appendix \Cref{app:images_framework}, for SDXL-Turbo, many images are visually corrupted and for Flux, the model fails to maintain prompt adherence.
For example, on the prompt ``a glass of orange juice with an orange peel stuck on the rim'', CMA-MAE generates images of dogs.
This indicates that its recorded diversity stems from corrupted or irrelevant outputs rather than genuine semantic variation, which alongside its low solution count (e.g., $5.92$ on Flux) explains its low Total Fitness scores.

The presence of visual artefacts is also evident in the fitness-maximising algorithms, Adam and Sep-CMA-ES.
A qualitative examination of images shows that both methods show signs of ``reward hacking" by introducing inconsequential artefacts to artificially inflate fitness scores.
In contrast, our approach discovers distinct interpretations such as varying viewing angles, background contexts, and compositions,  while maintaining high prompt alignment.
This is confirmed by the quantitative metrics: Adam (evaluated only on SDXL-Turbo) achieves the highest Total Fitness ($23.71$, $p < 3 \times 10^{-9}$) but suffers from a low Diversity score ($0.15$). 
Similarly, Sep-CMA-ES attains a Total Fitness comparable to our method on both SDXL-Turbo ($20.3$ vs $20.8$, $p > 0.35$) and Flux ($18.7$ vs $19.5$, $p>0.6$).
However, its diversity remains significantly lower on both ($0.20$ vs $0.47$, $p < 4 \times 10^{-10}$, for SDXL-Turbo, $0.095$ vs $0.472$, $p<6\times10^{-3}$ for Flux).
These results highlight that pure fitness maximisation often stagnates in local optima, converging to a single mode of the design space regardless of the base model used.

Finally, compared to QDHF (evaluated only on SDXL-Turbo), our method achieves significantly higher fitness ($16.1$ vs $20.8$, $p < 4 \times 10^{-7}$) and diversity ($0.28$ vs $0.47$, $p < 4 \times 10^{-8}$).
This suggests that projecting high-dimensional embeddings into lower-dimensional spaces may result in information loss and limit the algorithm's ability to retain diverse, high-performing solutions.

\subsection{Variation Operator Results}\label{sec:variation_results}

We next isolate the contribution of the $h$-space Spectral Mutation operator.
\Cref{tab:fid_operators} confirms that our method achieves the lowest FID score compared to all other variation operators on both SDXL-Turbo and Flux, confirming its efficacy at balancing quality and diversity.

\begin{table}[h!]
\centering
\tiny
\setlength{\tabcolsep}{2pt}
\renewcommand{\arraystretch}{1.2}
\begin{minipage}[t]{0.4\textwidth}
\centering
\caption{FID Score.}
\label{tab:fid_operators}
\vspace{-10pt}
\begin{tabularx}{\linewidth}{|X|c|c|}
\hline
\textbf{Algorithm} & \textbf{SDXL-Turbo} & \textbf{Flux}\\
\hline
Spectral-Guided (Ours) & \textbf{81.90} & \textbf{137.58}\\
\hline
Random Seeds & 100.12 &  180.03 \\
\hline
Diverse Diffusion & 104.59 & 185.94 \\
\hline
Particle Guidance & - & 186.74 \\
\hline
ImageBreeder & 207.75 & 247.03 \\
\hline
Gaussian & 93.34 & 144.14\\
\hline
Isoline-DD & 91.52 & 146.65 \\
\hline
Intermediate Gaussian & 109.73 & 163.31 \\
\hline
\end{tabularx}
\end{minipage}
\hfill
\begin{minipage}[t]{0.4\textwidth}
\centering
\caption{Number of Solutions.}
\label{tab:variation_number_sols}
\vspace{-10pt}
\begin{tabularx}{\linewidth}{|X|c|c|}
\hline
\textbf{Algorithm} & \textbf{SDXL-Turbo} & \textbf{Flux} \\
\hline
Spectral-Guided (Ours) &  \textbf{62.7 $\pm$ 8.0} & \textbf{58.0 $\pm$ 17}\\
\hline
Random Seeds &  \textbf{58.2 $\pm$ 15.0} &\textbf{54.0 $\pm$ 23} \\
\hline 
Diverse Diffusion & \textbf{58.1 $\pm$ 15.0} & \textbf{54.0 $\pm$ 23}\\
\hline
Particle Guidance & - & \textbf{50.0 $\pm$ 25}\\
\hline
ImageBreeder & 53.3 $\pm$ 20.2 & \textbf{54.0 $\pm$ 23}\\
\hline
Gaussian & \textbf{58.7 $\pm$ 15.3} & \textbf{54.0 $\pm$ 23}\\
\hline
Isoline-DD & \textbf{58.7 $\pm$ 15.3} & \textbf{54.0 $\pm$ 23}\\
\hline
Intermediate Gaussian & \textbf{60.0 $\pm$ 13.5} & \textbf{59.0 $\pm$ 17}\\
\hline
\end{tabularx}
\end{minipage}
\end{table}

\begin{figure}[t]
  \centering
  \subfloat[SDXL-Turbo.\label{fig:operator_pareto_turbo}]{%
    \includegraphics[width=0.4\textwidth]{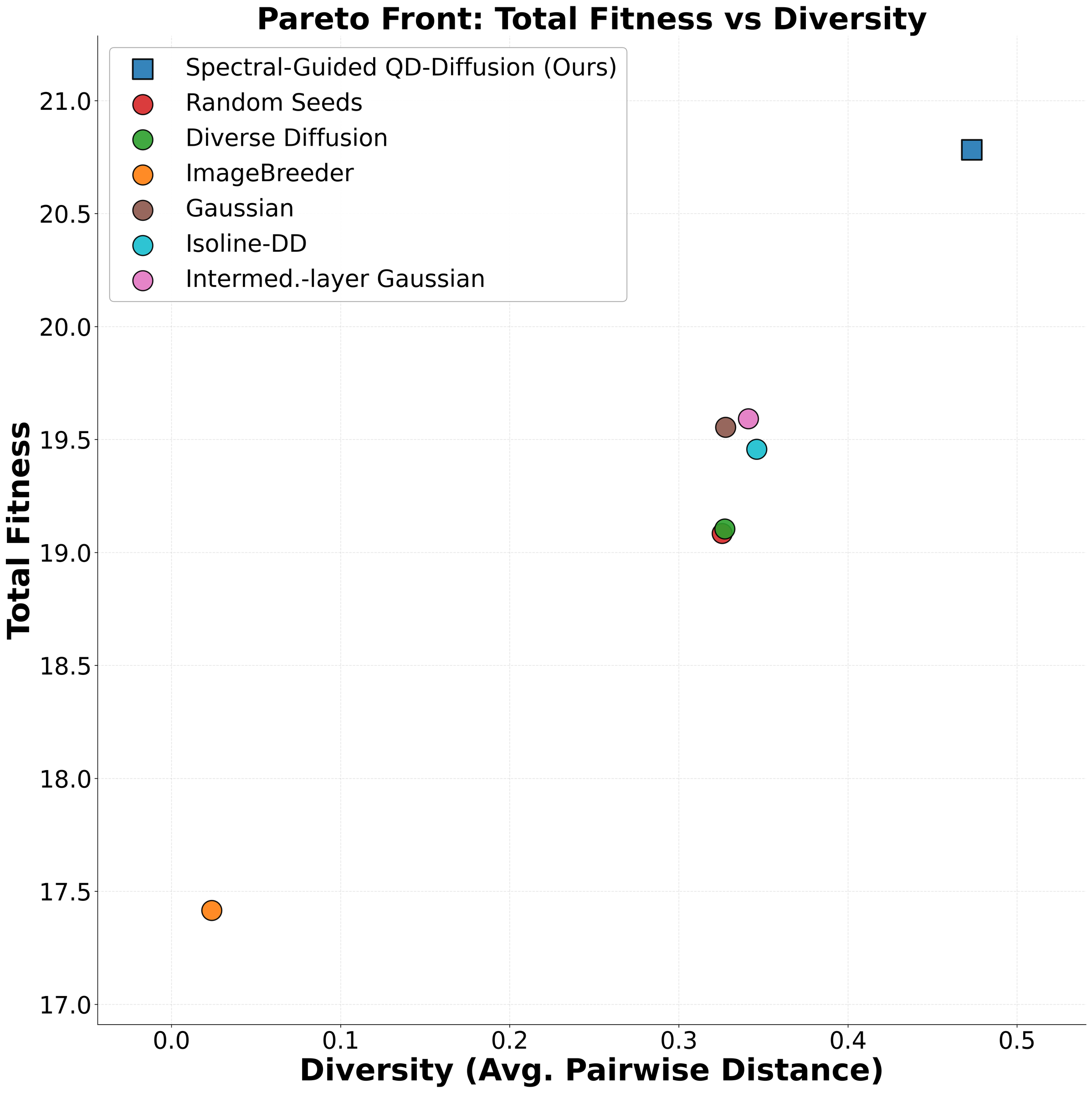}}
  \hspace{2em}
  \subfloat[Flux.\label{fig:operator_pareto_flux}]{%
    \includegraphics[width=0.4\textwidth]{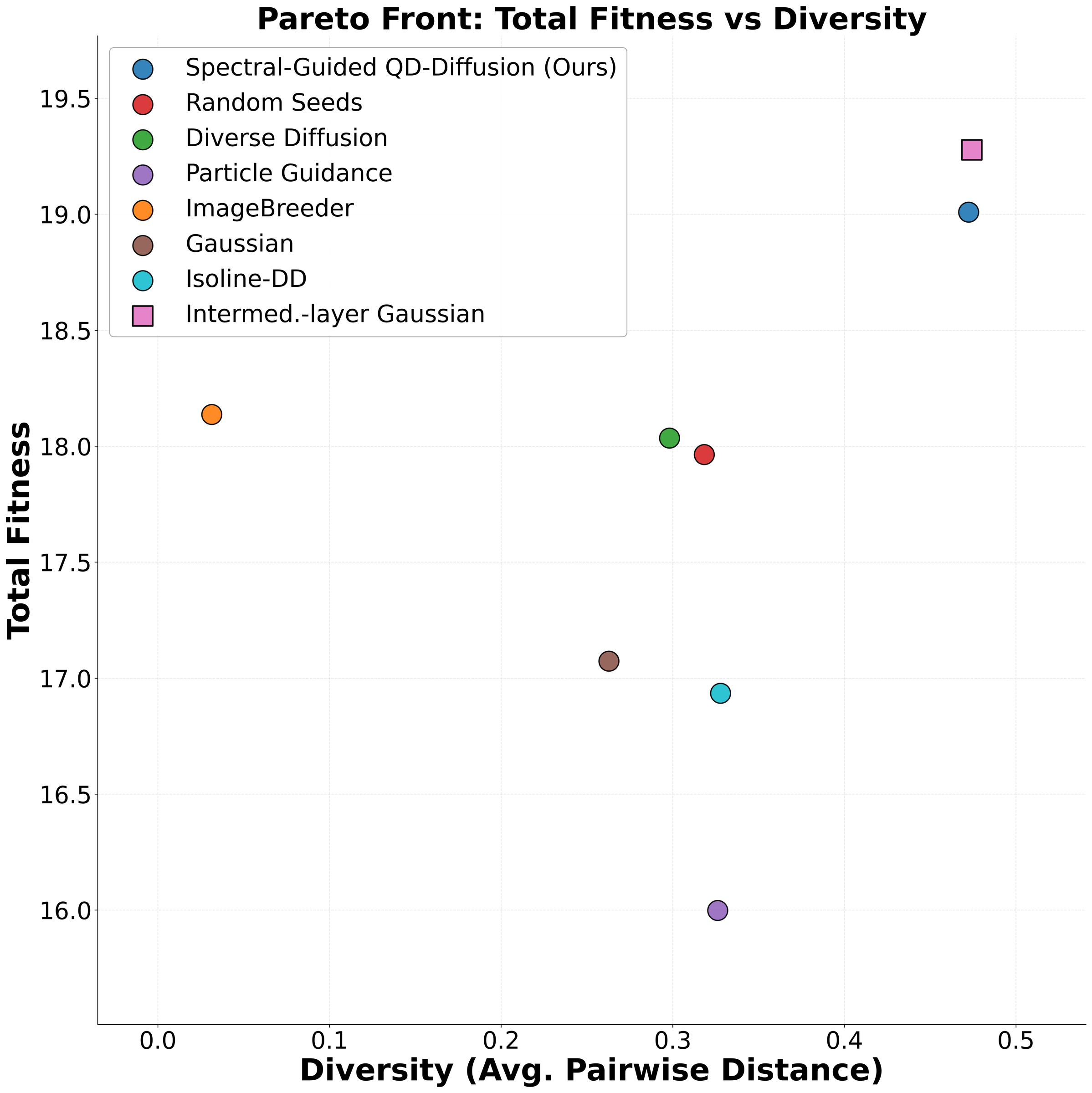}}
  \caption{Quality vs Diversity.}
  \label{fig:operator_pareto}
\end{figure}

\Cref{fig:operator_pareto} shows that on both SDXL-Turbo and Flux, our method Pareto-dominates all operators across both models except the Intermediate-Gaussian on Flux.
On Flux, the difference is statistically significant for both metrics for the Random Seeds and Imagebreeder baselines.
Otherwise, for all other baselines on both models, the difference in fitness is not statistically significant for the fitness metric ($p>0.05$), but it is for the diversity metric.
This confirms that traditional variation operators are inefficient at traversing high-dimensional latent spaces compared to our spectral approach.

Finally, we compare our approach to the Intermediate-layer Gaussian baseline.
Our method achieves significantly higher fitness on SDXL-Turbo ($20.78$ vs $19.59$, $p<5\times10^{-5}$) and diversity ($0.473$ vs $0.341$, $p<3\times10^{-10}$).
On Flux, the Intermediate-layer Gaussian records a higher diversity score ($0.498$ vs $0.489$) and comparable Total Fitness ($19.46$ vs $19.85$), but the difference is not significant in either case ($p>0.7$ for both).
Moreover, a qualitative examination in Appendix \Cref{app:images_variation} reveals that for both models, the Intermediate-layer Gaussian produces heavily corrupted images.
This indicates that its diversity arises from visual artefacts rather than genuine semantic variety, reinforcing the strength of the directional guidance provided by our $h$-layer spectral operator.

\section{Conclusion and Future Works}\label{sec:conclusion}
\vspace{-3mm}

In this work, we introduced \textit{Spectral-Guided Quality-Diversity Diffusion}, a framework designed to repurpose pre-trained diffusion models as searchable landscapes for evolutionary optimisation.
By exploiting the spectral properties of the generative prior through our novel $h$-space Spectral Mutation operator, our method effectively navigates high-dimensional latent manifolds to uncover diverse, high-fidelity portfolios of solutions.
Our empirical validation was conducted on the domain of text-to-image generation, where our method consistently outperforms $12$ baselines across two architectures in terms of FID score, demonstrating a superior balance of image quality and diversity.
While the framework was designed to be domain-agnostic, this cross-domain generalisability remains to be validated empirically.
Future work will therefore look to extend this framework beyond images to scientific discovery, specifically applying spectral mutations to biological sequence design.

\section*{AI Assistance Statement}
In this work, we used generative AI tools to aid and polish writing, for retrieval and discovery of related literature, to support research ideation and execution, and to draft sections of the paper text. We have not used generative AI tools to generate synthetic datasets or to prove mathematical claims; these are not applicable to this work.

We have reviewed all AI-assisted work. For example, when using AI to suggest related work, we read the corresponding literature to verify its relevance and manually added the references to our bibliography. Any drafts and edits of the text from AI was proofread for accuracy. We take responsibility for the final content of this work, including text, claims, and artifacts produced with the aid of generative AI.

\appendix
\numberwithin{equation}{section}

\bibliographystyle{abbrvnat} %
\bibliography{bibliography}

\section*{Supplementary Materials}
\section{Pseudocode}\label{app:pseudo}

\subsection{MAP-Elites Pseudocode}\label{app:me_pseudo}

\begin{algorithm}[H]
   \caption{MAP-Elites}
   \label{alg:me_pseudo}
\begin{algorithmic}
   \STATE {\bfseries Input:}  Search space $\mathcal{X}$, Feature function $\phi$, Fitness function $f$, Archive $\mathcal{A}$ with empty cells corresponding to discretised features
   \STATE Initialize archive $\mathcal{A}$ with random solutions sampled from $\mathcal{X}$
   \WHILE{not converged}
       \STATE Select parent $x$ uniformly at random from $\mathcal{A}$
       \STATE Generate offspring $x' \leftarrow \text{Mutate}(x)$
       \STATE Evaluate fitness $f(x')$ and feature $\phi(x')$
       \STATE Identify cell $c$ corresponding to $\phi(x')$
       \IF{cell $c$ is empty \textbf{or} $f(x') > f(\mathcal{A}[c])$}
           \STATE $\mathcal{A}[c] \leftarrow x'$ 
       \ENDIF
   \ENDWHILE
   \STATE {\bfseries Return} Archive $\mathcal{A}$
\end{algorithmic}
\end{algorithm}

\subsection{DNS Pseudocode}\label{app:dns_pseudo}

\begin{algorithm}[H]
   \caption{Dominated Novelty Search (DNS)}
   \label{alg:dns_pseudo}
\begin{algorithmic}
   \STATE {\bfseries Input:} Search space $\mathcal{X}$, Feature function $\phi$, Fitness function $f$, Population size $N$, Neighbours $k$
   \STATE Initialize population $P$ with $N$ random solutions sampled from $\mathcal{X}$
   \STATE Evaluate fitness $f(x)$ and feature $\phi(x)$ for all $x \in P$
   \WHILE{not converged}
       \STATE $Q \leftarrow \emptyset$
       \FOR{$x \in P$}
           \STATE $x' \leftarrow \text{Mutate}(x)$
           \STATE Evaluate $f(x')$ and $\phi(x')$
           \STATE $Q \leftarrow Q \cup \{x'\}$
       \ENDFOR
       \STATE $R \leftarrow P \cup Q$
       \FOR{$x \in R$}
           \STATE Find $k$-nearest neighbours of $x$ in $R$ with fitness $> f(x)$
           \STATE $\text{Score}(x) \leftarrow$ mean distance to these neighbours
       \ENDFOR
       \STATE $P \leftarrow$ Select top $N$ solutions from $R$ based on $\text{Score}$
   \ENDWHILE
   \STATE {\bfseries Return} Population $P$
\end{algorithmic}
\end{algorithm}

\section{Related Works}\label{app:related}

\subsection{Evolutionary Algorithms in Generative Design}
Several works have demonstrated that Evolutionary Algorithms can outperform gradient-based methods like Adam \citep{adam} when optimising the latent spaces of generative models such as GANs \citep{clare2023generating, fernandes2020evolutionary, costa2023exploring, roziere2020evolgan, innovation_engines}.
This observation extends to diffusion models, where EAs have been shown to locate higher-quality solutions than gradient descent \citep{neto2025evolutionary, Salvenmoser2025evolving}.
However, the majority of these works focus on single-objective fitness maximisation rather than explicitly optimising for diversity via Quality-Diversity (QD) algorithms.

While some approaches do search for diverse imagery, they often operate under significant constraints.
For instance, \citet{vinhas2016fitness} achieve diversity using Context Free Grammars, but their reliance on an unlimited archive classifies the method as novelty search rather than QD, and it does not operate within the high-dimensional embedding spaces typical of modern generative models.
Similarly, \citet{Salvenmoser2025evolving} employ an island-based Genetic Algorithm for open-ended search; however, this approach necessitates pre-defined artist styles per island, rendering it less generalisable.

Additionally, most of the existing works on using EAs for latent-space optimisation rely on traditional genetic operators, such as isotropic Gaussian noise.
An exception is ImageBreeder \citep{imagebreeder}, which proposes specialized crossover and mutation operators for images.
These include latent blending, channel-wise cuts, and mixed spatial cuts, all designed to recombine latent noise vectors from multiple parents.
However, these operators remain largely agnostic to the geometry of the generative model. 
Alternatively, while some methods evolve text prompts directly \citep{evolvetoinspire, simon_lluminate}, we avoid this to maintain a domain-agnostic framework applicable beyond language-conditioned tasks.

\subsection{Quality-Diversity for Generative Design}

Quality-Diversity algorithms have achieved notable success in illuminating the latent spaces of generative models.
Early applications coupled QD with GANs to generate diverse content, such as video game levels (e.g., Mario scenes) \citep{lsi} or face variations \citep{cma-mae, dqd}.
A limitation of these works is their reliance on hand-defined, low-dimensional features—for example, the ``number of enemies" in a level or ``hair length" in a face.
Our approach, by contrast, utilises a high-dimensional feature embedding space, ensuring generalisability to domains like biological sequence design where such semantic features are not easily defined.
Additionally, while methods like CMA-MAE \citep{cma-mae} leverage gradients of the fitness or feature to accelerate search, our method exploits the spectral properties of the generative model itself to guide exploration.

\subsection{Quality-Diversity for High-Dimensional Feature Spaces}
Scaling QD to the high-dimensional outputs of modern generative models remains non-trivial. The standard solution for high-dimensional features is CVT-ME \citep{cvt}, which uses Centroidal Voronoi Tessellations to discretise the feature space.
However, this approach suffers from the curse of dimensionality, leaving many cells empty in high-dimensional grids.
Dominated Novelty Search (DNS) \citep{dns} addresses this by maintaining an unstructured archive.
While effective in robotics tasks where offspring solutions remain in-distribution, we find that DNS requires modification for generative models to avoid retaining ``hallucinated" out-of-distribution artifacts.
Our method builds upon DNS by introducing a fitness thresholding mechanism to mitigate this issue.

Other approaches circumvent high-dimensional spaces by either a) using hand-defined metrics for diversity or b) incorporating human or LLM feedback to assess diversity \citep{simon_lluminate, deep_interactive_evolution}.
Notably, Quality-Diversity via Human Feedback (QDHF) \citep{qdhf} learns a projection layer (via contrastive loss from human/LLM feedback) to map high-dimensional embeddings into a manageable low-dimensional grid.
While effective, we demonstrate that directly optimising within the high-dimensional embedding space reduces the information loss associated with projection, leading to superior archive quality and diversity.

\subsection{Image Editing in Diffusion Models}

To design a more effective mutation operator, we draw inspiration from diffusion-based image editing.
Many works in this field use gradient-based optimization to maximize objectives like aesthetics \citep{deckers2023manipulating} or to alter an image towards a target text while preserving style \citep{kwon2023improvingdiffusionbasedimagetranslation, tumanyan2023plug}. However, in contrast to our approach, these methods typically optimize for a specific ``reward" or target, rather than exploring the space open-endedly.

Foundational analysis of the diffusion bottleneck, or $h$-space, has revealed that internal representations encode rich semantic geometry \citep{kwon2022diffusion, riemannian_geometry_latent_space}.
Our mutation operator is heavily inspired by work analysing the Riemannian Geometry of the latent space of diffusion models \citet{riemannian_geometry_latent_space}.
This work demonstrated that semantic edits can be generated by utilising a pullback metric to analyse the latent geometry of the latent-space. %
While this approach was developed for single-image \textit{editing}, we repurpose these geometric insights (specifically the spectral analysis of the Jacobian) as a mutation operator for population-based search.
Moreover, we extend this analysis beyond the latent noise vector to encompass all inputs of the diffusion model (including conditioning embeddings) considering them all as a genotype. 
Finally, we note the relevance of techniques such as Prompt-to-Prompt \citep{hertz2022prompt} and Null-Text Inversion \citep{mokady2023null} which established the importance of deterministic inversion mechanisms for preserving structure.
These mechanisms are fundamental to our approach; without the ability to deterministically invert the diffusion process, modifications made at semantically rich intermediate layers could not be mapped back to the initial noise space to form a stable, evolvable genotype.

\subsection{Image Diversity in Diffusion Models}

Beyond evolutionary strategies, several inference-time techniques have been proposed to mitigate the ``mode collapse" or lack of variety often observed in diffusion models.
Diverse Diffusion \citep{diversediffusion} addresses this by optimising the dispersion of initial latent noise vectors, utilising a rejection-sampling approach to ensure that a batch of generated images maintains a minimum pairwise distance in the latent space.
Similarly, Particle Guidance \citep{particle-guidance} replaces i.i.d. sampling with joint sampling of a batch of particles, augmenting the reverse diffusion process with the gradient of a pairwise-similarity potential (e.g. a kernel over image features) so that particles repel one another and spread out across the distribution.
While effective for batch-level variety, this approach is limited to the stochasticity of the prior and does not actively explore a semantic feature space.
Finally CADS (Condition-Annealed Diffusion Sampler) \citep{cads} is an inference-time, guidance-based diversity method that increases variety by annealing scheduled noise on the conditioning signal during sampling \citep{cads}.
CADS was not applied to SDXL-Turbo, as its gradual annealing of the conditioning signal requires a multi-step denoising trajectory that the few-step Turbo schedule does not provide; it was likewise excluded from Flux, whose guidance distillation eliminates the explicit conditional–unconditional guidance interaction that CADS depends on, rendering the method inapplicable in its intended form.

Unlike these methods, which focus on immediate batch variety, our archive-based approach cumulates semantic diversity over multiple iterations, allowing for a more thorough exploration of rare or distal feature regions.
Moreover, by operating on the genotype rather than the sampling trajectory, our method extends diversity-optimisation to one-step models such as SDXL-Turbo, where traditional guidance is inapplicable.

\section{Implementation Details}\label{app:implementations}

In this section, we detail the implementation specifics and hyperparameters used across all experiments. Our code is implemented using the Pyribs library \citep{pyribs} and is available at \codeurl.

\subsection{Global Hyperparameters}\label{app:global_hyperparams}

For SDXL-Turbo, all algorithms were allocated a total computational budget of 2000 evaluations. For Quality-Diversity (QD) algorithms, this corresponded to 125 iterations with a batch size of 16 offspring.

To fully explore the generative capabilities of SDXL-Turbo, we defined the genotype to include the latent noise vectors, the conditional prompt embeddings, and the pooled prompt embeddings. At each iteration, the batch of 16 offspring was divided among these components: 5 solutions mutated the latent vectors, 5 mutated the prompt embeddings, and 6 mutated the pooled prompt embeddings. The sole exception is the ImageBreeder baseline, which relies on operators designed specifically for image or latent space; for consistency in generalisation, we restricted ImageBreeder to mutate only the latent noise vector (see \Cref{app:baselines}).

For all archive-based algorithms (all algorithms except Adam and Sep-CMA-ES), the maximum archive capacity was set to 64 solutions.
The fitness threshold for Dominated Novelty Search (DNS) archives, including the passive archives used for reporting metrics on Adam and Sep-CMA-ES, was set to 0.3.

Due to the substantial computational cost associated with the Flux model, we restricted this evaluation to a reduced computational budget of 400 total evaluations, processed as 100 iterations with a batch size of 4.
The evaluation was performed on a subset of 12 prompts, consisting of one seed from each category of the Parti Prompts benchmark (prompts 1, 4, 7, etc. in Appendix \Cref{app:prompts_list}).

To mitigate runtime overhead, we applied 8-bit quantisation to the transformer component of the pipeline using the \texttt{BitsAndBytesConfig} from Hugging Face \citep{dettmers2022llm}.
Regarding the mutation operator, since Flux is a multi-step diffusion model, we randomly sampled the intermediate timestep $t_*$ from the interval $[0.2, 0.8]$ for the Jacobian calculation and we restricted the genotype to the latent embedding solely.
To accommodate the precision effects of quantisation, we adjusted the floating-point epsilon parameter for the Power Iteration SVD calculation to $1 \times 10^{-1}$.
All other hyperparameters—including the number of SVD iterations, the number of principal directions, and the fitness threshold—remained identical to the main experiments.
Crucially, we did not tune the mutation strength $\alpha$ for this architecture, testing the method's robustness to hyperparameter settings.

All experiments on both models were run on a high-performance computing cluster. Each experimental run was allocated a single node with 1 NVIDIA L40S GPU (48 GB VRAM), 16 CPU cores, and 48 GB of system memory.

\subsection{Spectral-Guided Diffusion Implementation}
\label{app:our_hyperparms}

Exact computation of the Jacobian is computationally intractable for high-dimensional diffusion models. We therefore approximate the principal singular vectors using the Power Iteration method described by \citet{riemannian_geometry_latent_space}. We estimate the first 3 principal directions using a maximum of 5 iterations and a floating-point epsilon of $1 \times 10^{-3}$.

During the mutation step, for each offspring, we randomly sample a direction $\mathbf{v}$ from these top 3 estimated singular vectors.
The genotype is then updated according to the rule $\mathbf{x}' = \mathbf{x} + \alpha \mathbf{v}$, where the step size $\alpha$ is sampled uniformly from the range $[-10, 10]$.
These parameters were consistent across all three genotype components and were used for any baseline employing the $h$-space Spectral Mutation Operator.

\subsection{Baseline Implementations}
\label{app:baselines}

\paragraph{CVT-ME}
As a grid-based algorithm, CVT-ME requires a defined discretisation of the feature space. We employed a Centroidal Voronoi Tessellation with 64 centroids to match the archive size of other methods. The bounds of the feature space were determined empirically to be $[-0.25, 0.25]$.
Due to the high computational cost of Flux, we did not evaluate CVT-ME on this model.
We instead prioritised CMA-MAE as the state-of-the-art grid-based baseline.
Since both methods utilise the same CVT archive structure, CMA-MAE provides a more state-of-the-art benchmark for archival performance within our limited computational budget.

\paragraph{Sep-CMA-ES}
We tuned the initial step size $\sigma$ for each genotype component independently across 5 random seeds, finding $\sigma = 0.1$ to be optimal for all components. We ran three independent Sep-CMA-ES processes for each genotype in parallel, corresponding to the batch sizes of the QD algorithms (5, 5, and 6).
Following the Pyribs implementation \citep{pyribs}, the number of parents selected at each iteration was set to half the batch size.

\paragraph{CMA-MAE}
Consistent with our other grid-based baselines, for CMA-MAE, we used a CVT archive with bounds of $[-0.25, 0.25]$. We employed the Improvement Ranker (ranker="imp") to prioritise solutions that discovered new cells or improved existing ones.
The initial step size was set to $\sigma_0=0.5$, and we utilised the basic restart rule to manage convergence.
We used all mutated all three genotypes in parallel, corresponding to the batch sizes of the QD algorithms (5, 5, and 6).
The archive learning rate was set to $0.01$ with a fitness threshold of $0.0$.

\paragraph{Adam}
We tuned the learning rate across 5 seeds and selected $\alpha = 0.01$. Unlike the population-based methods, Adam updates a single solution iteratively.
To maintain a comparable computational budget, we performed 2000 gradient steps.
Notably, Adam was not run on the Flux model, as the memory requirements for backpropagation on Flux exceeded available hardware limits, resulting in Out-Of-Memory (OOM) errors.

\paragraph{QDHF}
We followed the open-source implementation of Quality-Diversity via Human Feedback (QDHF).
The algorithm learns a projection layer mapping high-dimensional CLIP embeddings to a low-dimensional latent space.
We set this latent dimension to 2 and trained the projection layer at iterations corresponding to $1\%$, $10\%$, $25\%$, and $50\%$ of the total budget.
Training used a minibatch size of 32 and a buffer of 1000 preference data points.
For the initial warmup phase (iteration 0), where no solutions exist, we generated 200 random images with a latent sigma of 0.01.
These 200 image evaluations was given in addition to the main algorithmic 2000 evaluations budget.
We excluded QDHF from the Flux experiments because the 200 "free" warmup images required for Flux would constitute half of the main method’s total computational budget, making the comparison prohibitively expensive.

\paragraph{Diverse Diffusion}
We implemented the ``Max" variant of Diverse Diffusion \citep{diversediffusion}, adapted to operate as an emitter within our experimental framework.
For each slot in the batch, the algorithm samples $n_{\text{max}}=10$ candidate latent vectors from a standard normal distribution and selects the one that maximises the minimum $L_2$ distance to all previously selected vectors in the current batch.
Following the original implementation, we used the latent embeddings as genotypes.

\paragraph{Particle Guidance}
We implemented Particle Guidance (PG) \citep{particle-guidance}, generating each batch of particles jointly so that they repel one another throughout denoising. Particles were initialised from independent standard-normal noise and shared the parent's prompt and pooled embeddings.
Following the original work, we induced repulsion with an RBF kernel in latent space, using the median heuristic for the bandwidth and scaling the repulsive gradient by a coefficient of $0.5$ and the current noise level; repulsion was disabled once $\sigma_t < 0.5$ to preserve detail near the end of sampling. We used $28$ denoising steps and an embedded guidance scale of $3.5$.
Particle Guidance was not applied to SDXL-Turbo, as its repulsion mechanism requires the samples to interact across multiple denoising steps, which the few-step Turbo schedule does not provide.

\paragraph{ImageBreeder}
ImageBreeder introduces nine mutation operators for diffusion models. We utilised the `mixed-all' operator strategy, which randomly samples an operator at each iteration.
To ensure domain agnosticism, we excluded the pixel-level operators (`pixel-blend' and `pixel-blend-with-score') and utilised only the latent-based operators. 
Unlike other baselines initialized with a single seed, ImageBreeder relies on crossover operations which require population diversity to function effectively.
Consequently, to avoid bias from a stagnant initial population, we seeded ImageBreeder with two distinct random images.

\paragraph{Gaussian Noise Mutation}
We tuned the mutation strength ($\sigma$) for each genotype component separately over 5 seeds. The optimal parameters were found to be $\sigma=0.1$ for latent vectors, and $\sigma=0.4$ for both prompt and pooled prompt embeddings.

\paragraph{IsoLine-DD Variation Operator}
For the IsoLine operator, we used the same isotropic $\sigma_1$ values as the Gaussian Noise baseline. We tuned the line mutation strength ($\sigma_2$) and found $\sigma_2=0.1$ to be optimal for all genotype components.

\paragraph{Intermediate Layer Gaussian}
To isolate the effect of spectral guidance, this baseline applies Gaussian noise at the intermediate $h$-layer rather than using singular vectors. Tuning over 5 seeds yielded an optimal $\sigma=0.4$ for all three genotype components.

\subsection{Baselines Evaluation}
\label{app:baselines-evaluation}
We evaluate QD baseline algorithms by computing the evaluation metrics established in \Cref{sec:setup} directly on the solutions in their final archives.
For non-QD algorithms (i.e., CMA-ES, Adam, Random Seeds), we maintain a passive DNS archive with thresholding to calculate metrics comparably.
This passive archive collects the solutions generated by the algorithm at each generation and is used exclusively for metrics reporting; it does not influence the main algorithm's search process.
To ensure that performance differences arise from algorithmic dynamics every algorithm was seeded with the identical starting image for each prompt.

\subsection{Parti Prompts Distribution}\label{app:prompts_analysis}

\begin{figure}[ht!]
    \centering
    \includegraphics[width=0.75\linewidth]{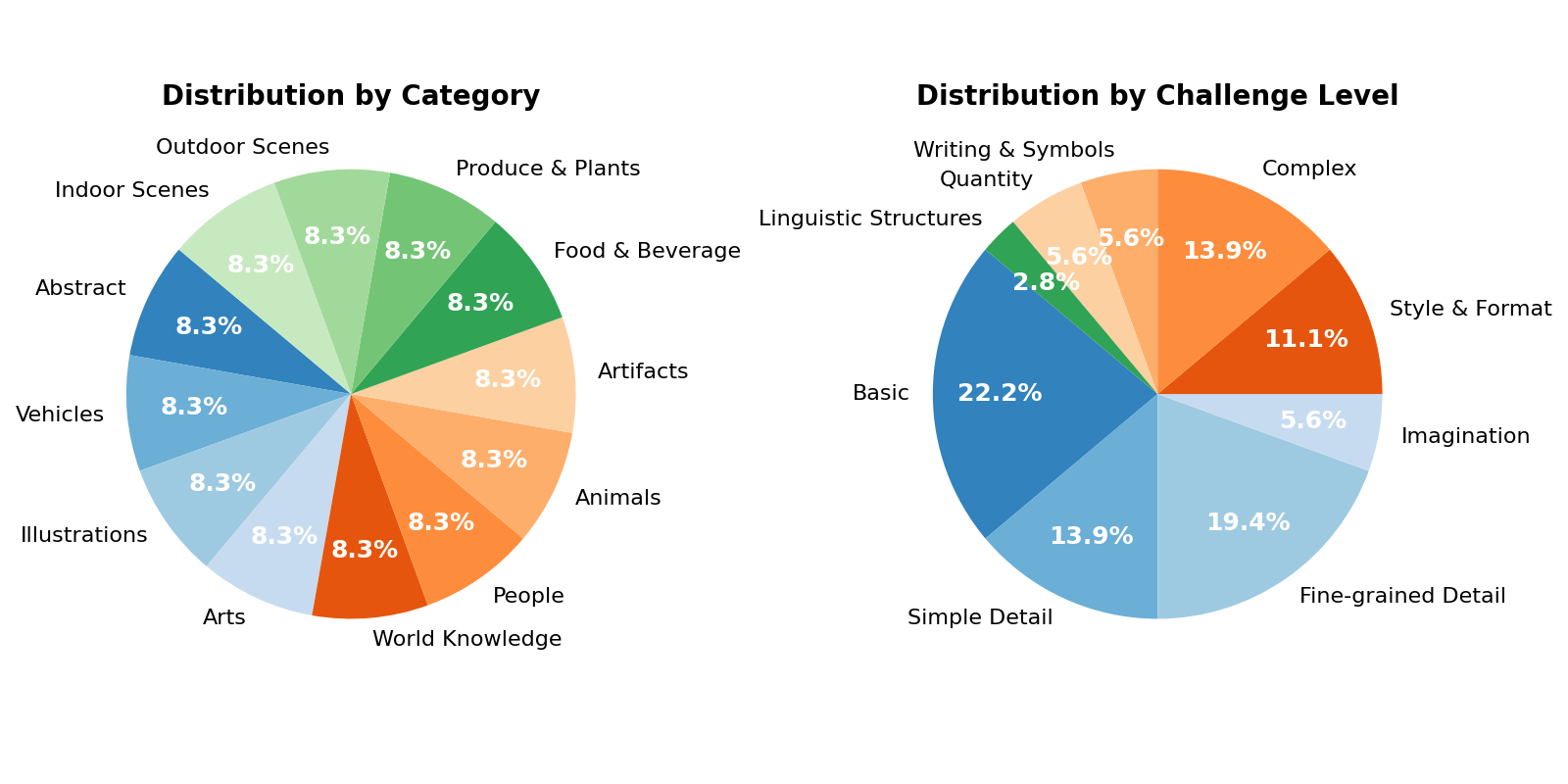}
    \caption{Pie charts representing the Category and Challenge Distribution of the sampled prompts.}
    \label{fig:prompts_distribution}
\end{figure}

\Cref{fig:prompts_distribution} shows the Category and Challenge Distribution of prompts selected randomly from the Parti Prompts Dataset. We sampled 3 prompts from each of the categories leading to a total of 36 prompts (see Appendix \Cref{app:prompts_list} for the full list). The most common Challenge levels were ``Basic" and ``Fine-grained Detail".

\subsection{FID Metric Implementation}\label{app:fid_implementation}

To evaluate the visual quality and diversity of the generated archives, we calculate the Fréchet Inception Distance (FID) metric.
This metric is computed using high-level feature representations extracted from the pre-trained Inception-v3 network \citep{szegedy2016rethinking}, with the MS-COCO 2017 validation set serving as the reference distribution for real-world imagery \citep{lin2014microsoft}.

\subsubsection{FID Extrapolation}
Since the FID metric is sensitive to sample size, we aggregate the images from all experimental runs and seeds for each baseline into a single combined archive. 
This accumulation is performed specifically to maximize the available sample size, thereby reducing statistical variance and providing a more robust representation of each algorithm’s output distribution.
Despite this aggregation, the total archive sizes still vary between methods. We therefore adopt the $FID_\infty$ extrapolation technique proposed by \citet{fid_estimate} to ensure an unbiased comparison. 

Following the observation that the expected FID value is approximately linear with respect to the inverse of the number of samples $N$, we compute the raw FID at multiple subsampled intervals: $N \in \{200, 400, \dots, 2000\}$.
We then perform a linear least-squares regression of these values against $1/N$.
The reported FID is the intercept of this regression line as $N\rightarrow\infty$, providing an estimate of the model’s performance that is independent of the specific archive size.
Algorithms that failed to produce the minimum required sample size for the regression analysis (e.g., $N<200$) were excluded from the FID$_\infty$ calculation, as a sufficient number of data points is mathematically necessary to produce a stable linear fit and avoid singular covariance matrices in the high-dimensional Inception feature space.

\section{Ablation Experiments}\label{app:ablations}

In this section, we include ablation studies on key hyperparameters of our method.
Due to computational constraints, this study was performed only on SDXL-Turbo.
The fitness threshold removal (\Cref{app:no_threshold}) and genotype ablation (\Cref{app:genotype_ablation}) were assessed using all $36$ prompts, while the others used $12$ prompts, one from each category.

\subsection{Fitness Threshold Ablation}\label{app:threshold_ablation}

\subsubsection{No Fitness Threshold}\label{app:no_threshold}

To validate the necessity of the fitness threshold mechanism within our method, we conduct an ablation study where this component is removed.
Results for this comparison are organised in Table~\ref{tab:threshold_ablation_results}. 

\begin{table}[htbp]
\centering
\scriptsize
\setlength{\tabcolsep}{2pt} %
\renewcommand{\arraystretch}{1.2} %
\caption{Comparison between our full method and the fitness threshold ablation across diversity and fitness metrics.}
\label{tab:threshold_ablation_results}
\begin{tabularx}{0.9\columnwidth}{|X|c|c|c|}
\hline
\textbf{Algorithm} &  Total Fitness & Avg. Pairwise Dist. & Num. Solutions \\ \hline
Spectral-Guided QD-Diffusion (Ours) & \textbf{20.8 $\pm$ 3.0} &  0.473 $\pm$ 0.08 & \textbf{62.7 $\pm$ 8.0} \\ \hline
Fitness Threshold Ablation & 17.56 $\pm$ 1.8 &  \textbf{0.673 $\pm$ 0.07} & \textbf{64.0 $\pm$ 0.0} \\
\hline
\end{tabularx}
\end{table}

As expected, the results demonstrate that while the ablation variant achieves a significantly higher diversity score ($p < 2 \times 10^{-10}$), it suffers from a Total Fitness score much lower than our full method ($p < 1 \times 10^{-2}$).
This indicates an archive populated by solutions that are novel but ultimately invalid or semantically inconsistent.

\begin{figure}[ht]
    \centering
    \subfloat[Fitness Threshold Ablation]{
        \includegraphics[width=0.3\textwidth]{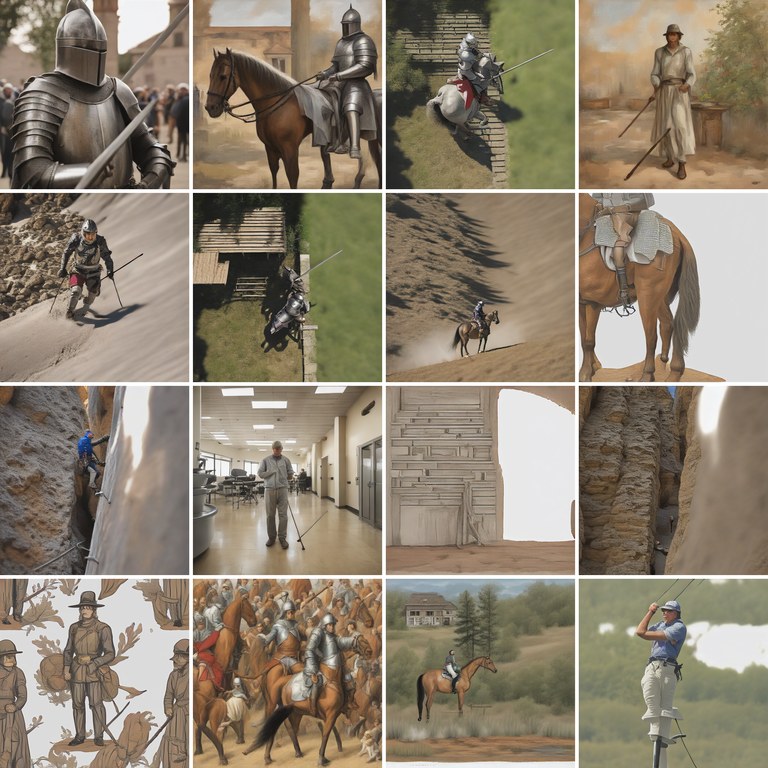}
        \label{fig:ablation_a}
    }
    \hspace{1.2cm}
    \subfloat[Spectral-Guided QD-Diffusion (Ours)]{
        \includegraphics[width=0.3\textwidth]{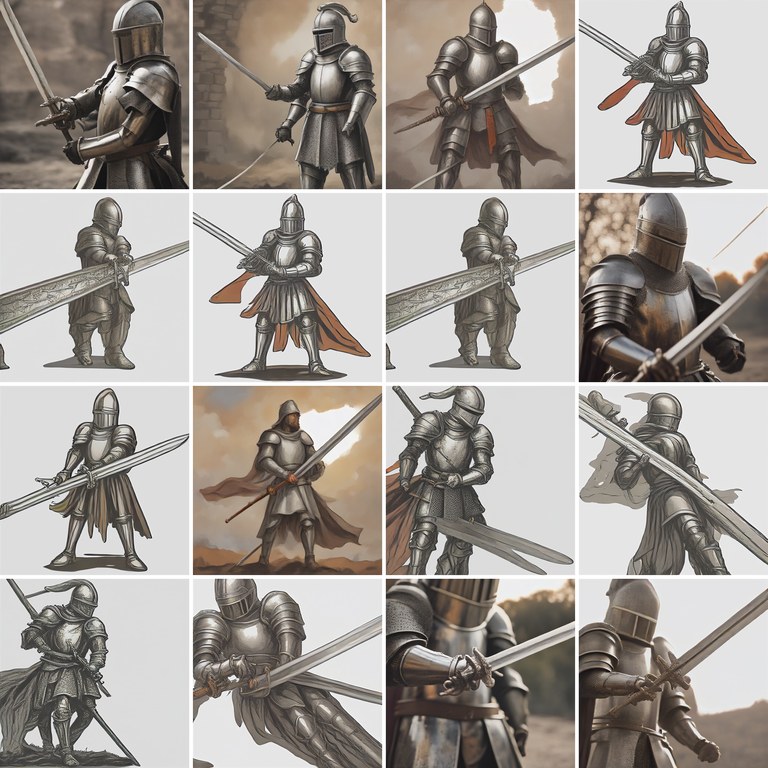}
        \label{fig:ablation_b}
    }
    \caption{Qualitative comparison for the prompt ``a knight holding a long sword''. The ablation (a) exhibits high diversity but often fails to meet semantic constraints, whereas our method (b) maintains high fitness and validity.}
    \label{fig:ablation_visuals}
\end{figure}

This disparity is further clarified through qualitative analysis. Figure~\ref{fig:ablation_visuals} displays the top-16 solutions for the prompt ``a knight holding a long sword'' generated by the ablation and our full method.
While some of the images produced by the ablation are novel and interesting, the archive contains many low-quality examples.
For instance, in one image the ``knight'' is rendered as an old man and his ``sword'' is a walking stick. 
Similar behaviour is observed for more complex prompts, such as ``a drop-top sports car coming around a bend'', where the ablation baseline tends to produce stretched car geometries, toy cars, or empty roads.
While these results highlight the raw creativity of the $h$-space Spectral Mutation operator, they also underscore the critical necessity of our fitness gating mechanism to maintain semantic integrity.

\subsubsection{Fitness Threshold Robustness}

Beyond the binary removal of the threshold, we evaluate the robustness of our method to specific threshold choices by testing a range of values. The results are summarised in Table~\ref{tab:threshold_robustness}.

\begin{table}[htbp]
\centering
\scriptsize
\setlength{\tabcolsep}{2pt} %
\renewcommand{\arraystretch}{1.2} %
\caption{Robustness analysis of the fitness threshold. Performance varies smoothly, with 0.3 serving as a stable default.}
\label{tab:threshold_robustness}
\begin{tabularx}{0.9\columnwidth}{|X|c|c|c|}
\hline
\textbf{Threshold Value} &  Total Fitness & Avg. Pairwise Dist. & Num. Solutions \\ \hline
0.26 & \textbf{20.090 $\pm$ 0.965} & \textbf{0.576 $\pm$ 0.069} & \textbf{64.000 $\pm$ 0.000} \\ \hline
0.28 & \textbf{20.673 $\pm$ 1.100} & 0.542 $\pm$ 0.068 & \textbf{64.000 $\pm$ 0.000} \\ \hline
0.30 (Default) & \textbf{21.104 $\pm$ 1.143} & 0.500 $\pm$ 0.072 & \textbf{64.000 $\pm$ 0.000} \\ \hline
0.32 & \textbf{17.647 $\pm$ 8.035} & 0.344 $\pm$ 0.146 & \textbf{52.000 $\pm$ 21.709} \\ \hline
0.34 & \textbf{12.454 $\pm$ 9.789} & 0.256 $\pm$ 0.137 & \textbf{36.083 $\pm$ 24.645} \\
\hline
\end{tabularx}
\end{table}

The system demonstrates significant robustness, with performance metrics varying smoothly across the tested range.
Lowering the threshold below 0.3 allows for a slight increase in diversity by permitting more marginal solutions, while raising the threshold above 0.3 results in a monotonic decline in archive coverage and Total Fitness.
This suggests that while the mechanism is robust, the default threshold effectively balances semantic integrity with diversity.

\subsection{Genotype Ablation}\label{app:genotype_ablation}

In this section, we analyse the impact of different genotype components on the algorithm's performance.
Our framework, as proposed in \Cref{sec:method_framework}, is designed to be general, allowing any input to the generative model to serve as a mutable genotype.
For generality, our main experiments employed a mixed strategy, mutating the latent noise, prompt embeddings, and pooled prompt embeddings in equal proportions.
To verify the validity of this choice, we conducted an ablation study isolating each component.
All other hyperparameters were identical to the main experiments (\Cref{app:global_hyperparams}).

\begin{table}[htbp]
\centering
\scriptsize %
\setlength{\tabcolsep}{2pt} %
\renewcommand{\arraystretch}{1.2} %
\caption{Performance comparison across algorithms. Bold values indicate the best performer or algorithms not significantly worse than the best (p $\geq$ 0.05).}
\label{tab:genotypes_ablation}
\begin{tabularx}{0.9\columnwidth}{|X|c|c|c|}
\hline
\textbf{Algorithm} & Total Fitness& Diversity& No. of Solutions \\
\hline
Mixed Genotype & \textbf{20.782 $\pm$ 3.013} & \textbf{0.473 $\pm$ 0.083}  & \textbf{62.667 $\pm$ 8.000} \\
\hline
Latent Noise &  \textbf{20.393 $\pm$ 4.238} & 0.351 $\pm$ 0.093 &  \textbf{61.333 $\pm$ 11.151} \\
\hline
Prompt Embedding & 19.988 $\pm$ 4.005 & \textbf{0.480 $\pm$ 0.096}  & \textbf{61.333 $\pm$ 11.151} \\
\hline
Pooled Embedding & 18.792 $\pm$ 6.042 & 0.357 $\pm$ 0.150 & \textbf{57.056 $\pm$ 16.805} \\
\hline
\end{tabularx}
\end{table}

\Cref{tab:genotypes_ablation} presents the results averaged across all 36 prompts.
We observe no statistically significant difference in the Number of Solutions across any of the methods, indicating all genotype strategies are capable of populating the archive. 
In terms of Total Fitness, the Mixed Genotype and Latent Noise approaches perform best, significantly outperforming the Pooled Prompt Embedding. However, regarding Diversity, both our Mixed Genotype strategy and the Prompt Embedding ablation achieve the highest scores, significantly surpassing the Latent Noise and Pooled Prompt Embedding approaches.

These results suggest a trade-off: mutating the Latent Noise is highly effective for maintaining fitness but yields lower diversity, while mutating the Prompt Embedding drives significant semantic variation (Diversity) at a slight cost to fitness.
Our Mixed Genotype approach successfully leverages the strengths of both, achieving top-tier performance in both Total Fitness and Diversity. Consequently, we retain the mixed strategy for our final experiments to ensure robustness and generalisability across different generative tasks.

\subsection{Mutation Strength (\texorpdfstring{$\alpha$}{alpha}) Ablation}\label{app:alpha_ablation}

We further investigate the sensitivity of our method to the mutation strength parameter, $\alpha$, which dictates the magnitude of the $h$-space perturbation.
Table~\ref{tab:alpha_ablation} compares our default value against halving and doubling the magnitude.

\begin{table}[htbp]
\centering
\scriptsize
\setlength{\tabcolsep}{2pt}
\renewcommand{\arraystretch}{1.2}
\caption{Ablation of mutation strength $\alpha$. The method remains stable across varying perturbation magnitudes.}
\label{tab:alpha_ablation}
\begin{tabularx}{0.9\columnwidth}{|X|c|c|c|}
\hline
\textbf{Algorithm} & Total Fitness & Avg. Pairwise Dist. & Num. Solutions \\ \hline
$\alpha = 5$ & \textbf{19.668 $\pm$ 5.020} & 0.451 $\pm$ 0.070 & \textbf{60.000 $\pm$ 13.856} \\ \hline
$\alpha = 10$ (Default) & \textbf{21.104 $\pm$ 1.143} & 0.500 $\pm$ 0.072 & \textbf{64.000 $\pm$ 0.000} \\ \hline
$\alpha = 20$ & \textbf{21.087 $\pm$ 1.015} & \textbf{0.562 $\pm$ 0.059} & \textbf{64.000 $\pm$ 0.000} \\ \hline
\end{tabularx}
\end{table}

The results indicate that doubling or halving the perturbation magnitude has a negligible effect on overall performance.
All configurations achieved comparable fitness and diversity, suggesting that the Spectral Mutation operator is not hyper-sensitive to the exact scale of $\alpha$.

\subsection{Spectral Direction (\texorpdfstring{$k$}{k}) Iteration Ablation}\label{app:svdk_ablation}

To determine the number of spectral directions required to effectively guide the mutation, we ablate the value of $k$ in the SVD-based operator.
Results are organised in Table~\ref{tab:svdk_results}.

\begin{table}[htbp]
\centering
\scriptsize
\setlength{\tabcolsep}{2pt}
\renewcommand{\arraystretch}{1.2}
\caption{Impact of the number of spectral directions $k$ on algorithm performance.}
\label{tab:svdk_results}
\begin{tabularx}{0.9\columnwidth}{|X|c|c|c|}
\hline
\textbf{Algorithm} & Total Fitness & Avg. Pairwise Dist. & Num. Solutions \\ \hline
$k = 1$ & \textbf{18.617 $\pm$ 6.630} & \textbf{0.453 $\pm$ 0.083} & \textbf{56.615 $\pm$ 18.026} \\ \hline
$k = 3$ (Default) & \textbf{21.104 $\pm$ 1.090} & \textbf{0.495 $\pm$ 0.071} & \textbf{64.000 $\pm$ 0.000} \\ \hline
$k = 5$ & \textbf{21.091 $\pm$ 1.104} & \textbf{0.480 $\pm$ 0.075} & \textbf{64.000 $\pm$ 0.000} \\ \hline
$k = 10$ & \textbf{21.191 $\pm$ 1.158} & \textbf{0.465 $\pm$ 0.068} & \textbf{64.000 $\pm$ 0.000} \\ \hline
\end{tabularx}
\end{table}

Performance effectively plateaus at $k=3$, with diminishing returns observed for higher values.
Only the reduction to $k=1$ leads to a meaningful drop in archive stability and fitness.
This suggests that a small number of spectral directions is sufficient to capture the semantically active subspace required for effective Quality-Diversity exploration.

\subsection{Targeted Hidden Layer Ablation}\label{app:layer_ablation}

Finally, we explore the impact of targeting different architectural blocks within the U-Net for spectral mutation.
Table~\ref{tab:layer_results} compares our default choice (\texttt{mid\_blocks}) against other potential targets.

\begin{table}[htbp]
\centering
\scriptsize
\setlength{\tabcolsep}{2pt}
\renewcommand{\arraystretch}{1.2}
\caption{Comparison of spectral mutation performance when applied to different U-Net layers.}
\label{tab:layer_results}
\begin{tabularx}{0.9\columnwidth}{|X|c|c|c|}
\hline
\textbf{Target Layer} & Total Fitness & Avg. Pairwise Dist. & Num. Solutions \\ \hline
\texttt{mid\_blocks} (Default) & \textbf{21.104 $\pm$ 1.143} & \textbf{0.500 $\pm$ 0.072} & \textbf{64.000 $\pm$ 0.000} \\ \hline
\texttt{down\_blocks.2} & \textbf{21.380 $\pm$ 1.170} & \textbf{0.499 $\pm$ 0.071} & \textbf{64.000 $\pm$ 0.000} \\ \hline
\texttt{up\_blocks.1} & \textbf{20.021 $\pm$ 5.551} & \textbf{0.465 $\pm$ 0.093}  & \textbf{59.636 $\pm$ 14.473} \\ \hline
\texttt{up\_blocks.2} & \textbf{20.098 $\pm$ 5.596} & \textbf{0.467 $\pm$ 0.109} & \textbf{59.636 $\pm$ 14.473} \\ \hline
\end{tabularx}
\end{table}

The choice of layer has a minimal overall impact on the results.
However, we observe that targeting decoder-side layers (\texttt{up\_blocks}) results in slightly lower stability and fitness compared to bottleneck or encoder-side layers, justifying our choice of the central blocks.

\section{Sample Image Results}

\subsection{Framework Experiment Images}\label{app:images_framework}

\begin{figure}[htbp]
    \centering
    \scalebox{0.9}{
        \begin{minipage}{\textwidth}
        
        \comparisonBlockSmall{baseline}
            {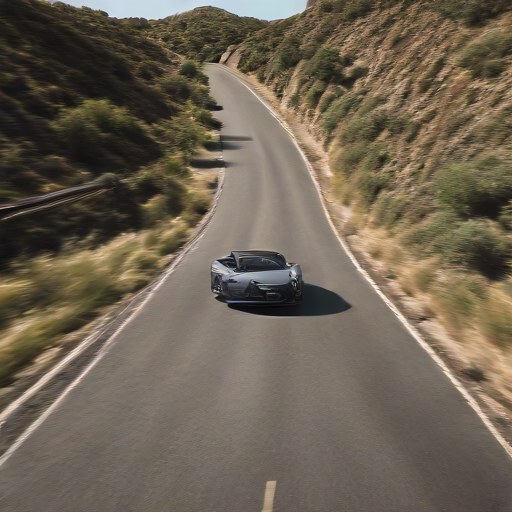}
            {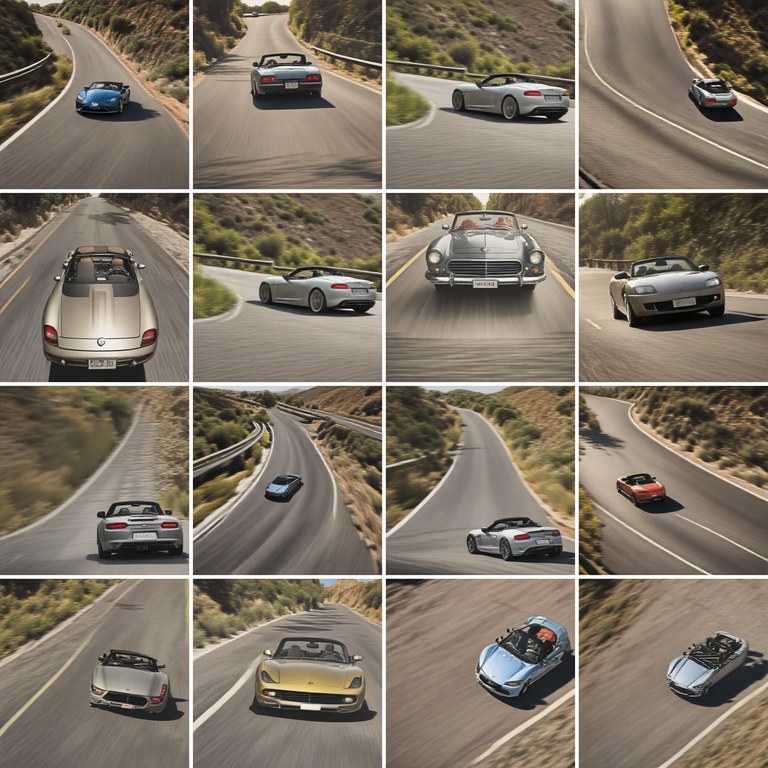}
            {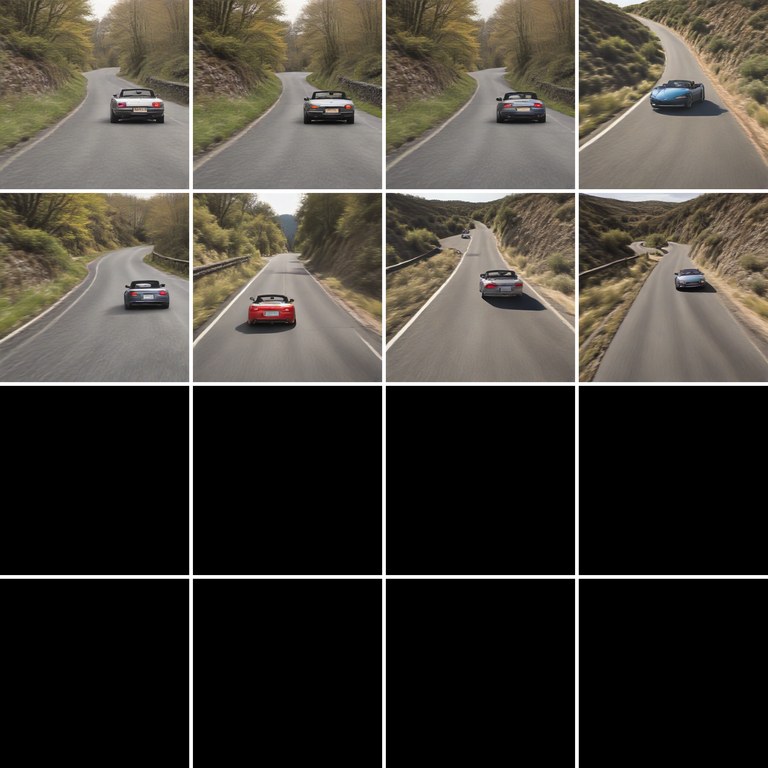}
            {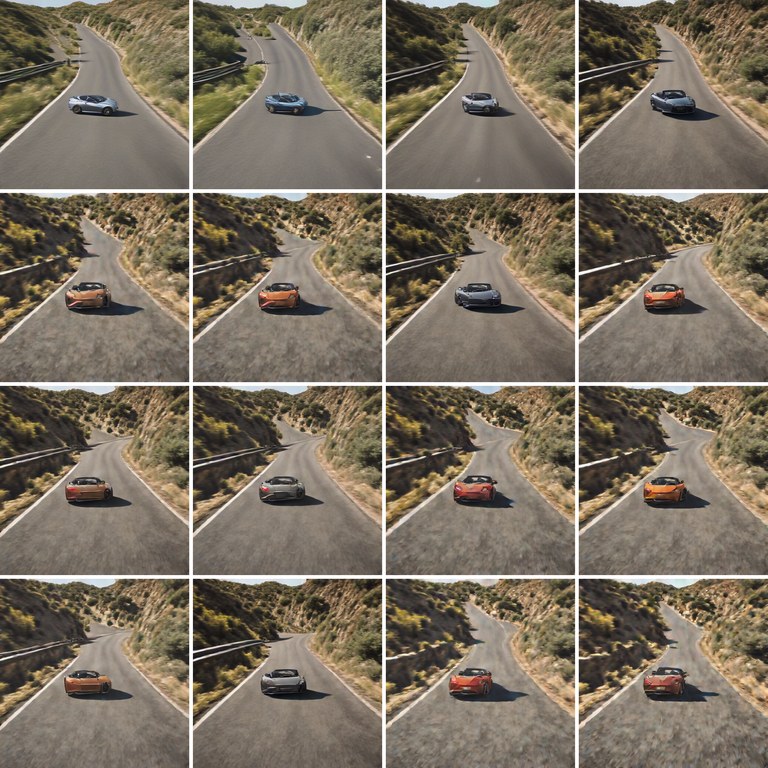}
            {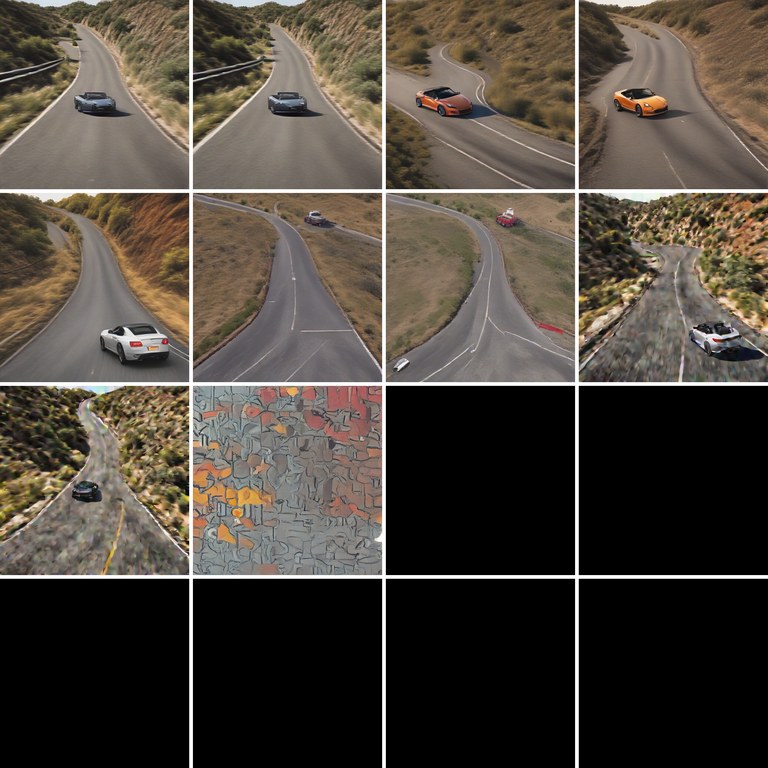}
            {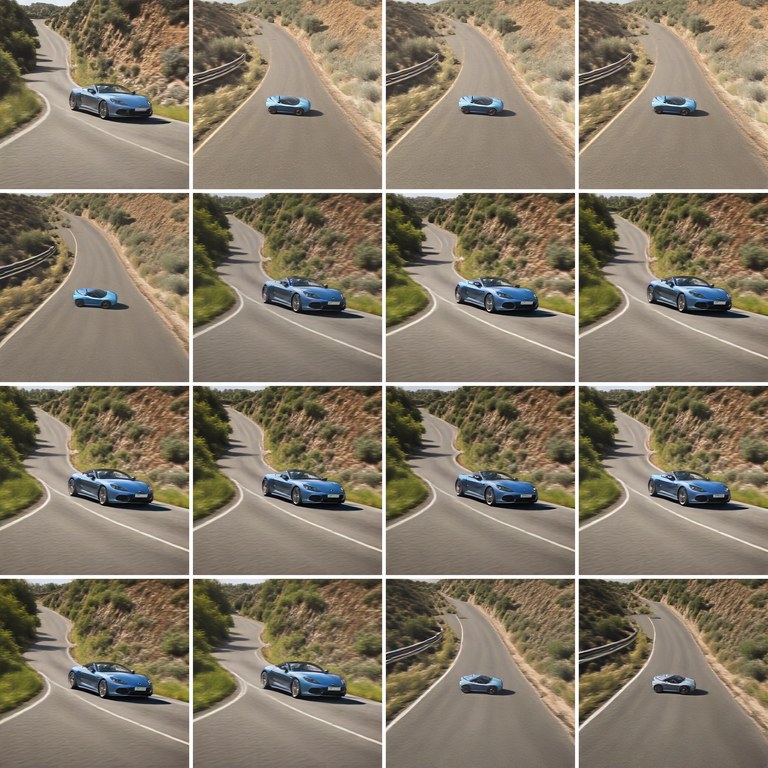}
            {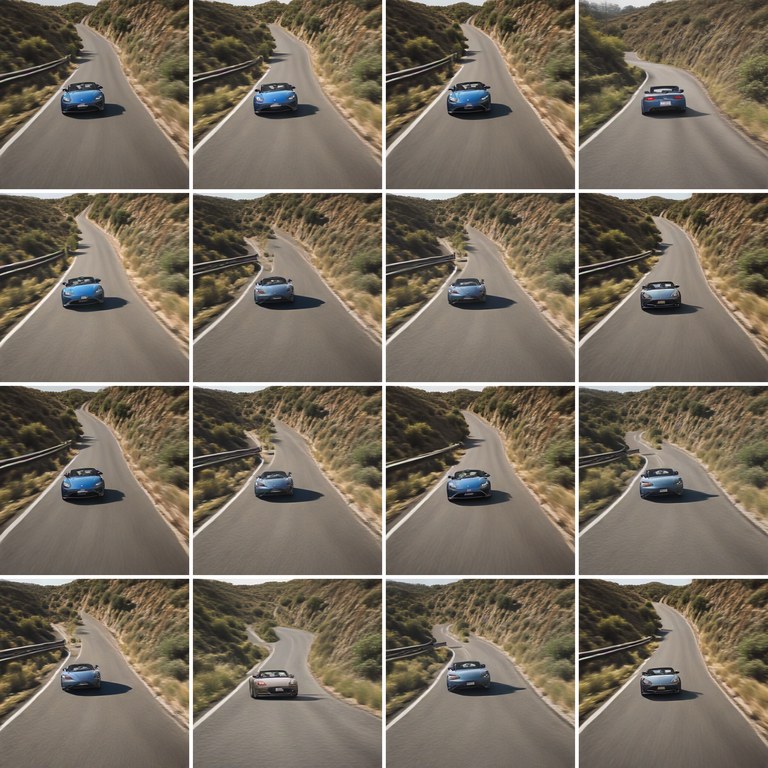}
            {a drop-top sports car coming around a bend in the road}

        \comparisonBlockSmall{baseline}
            {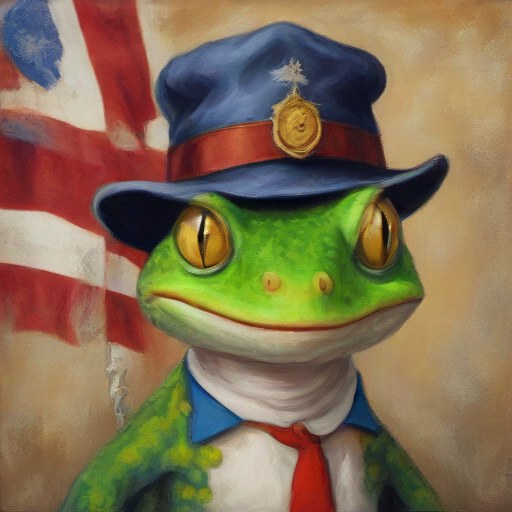}
            {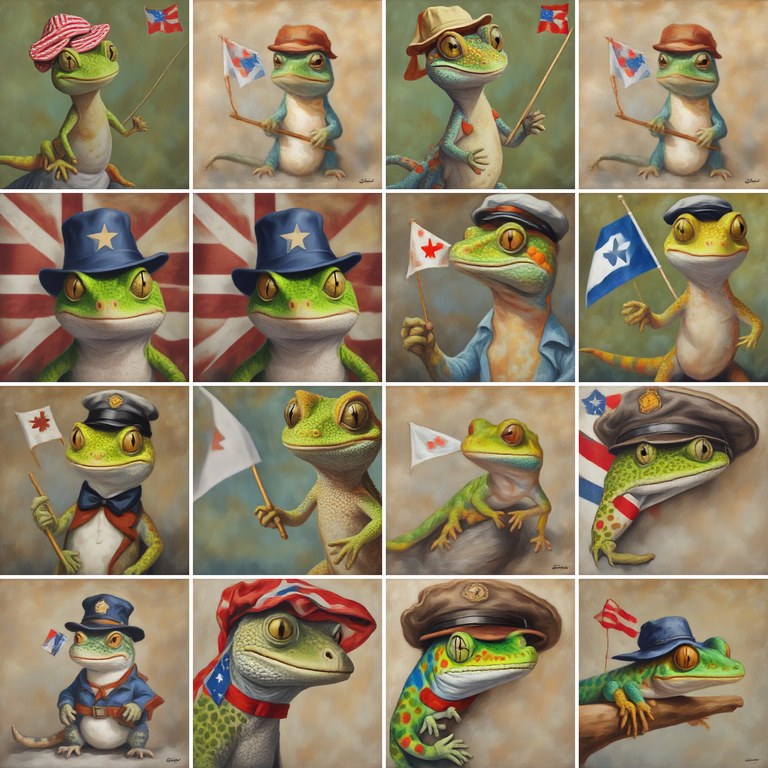}
            {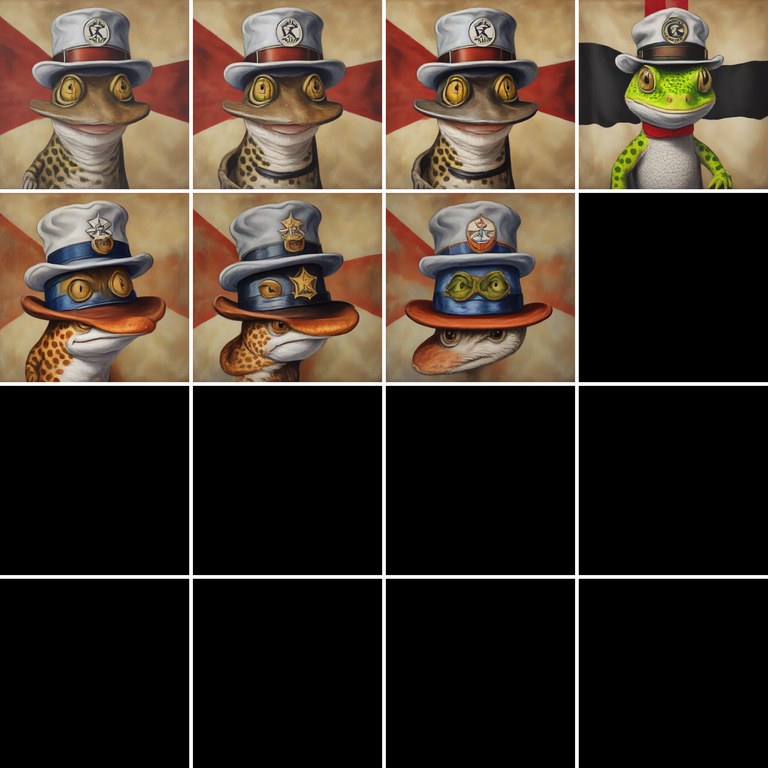}
            {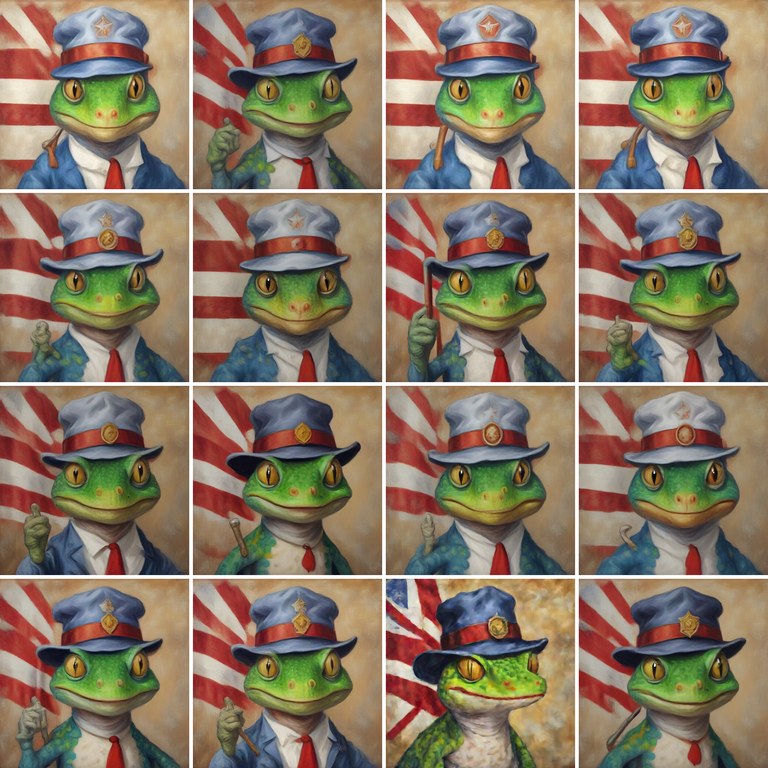}
            {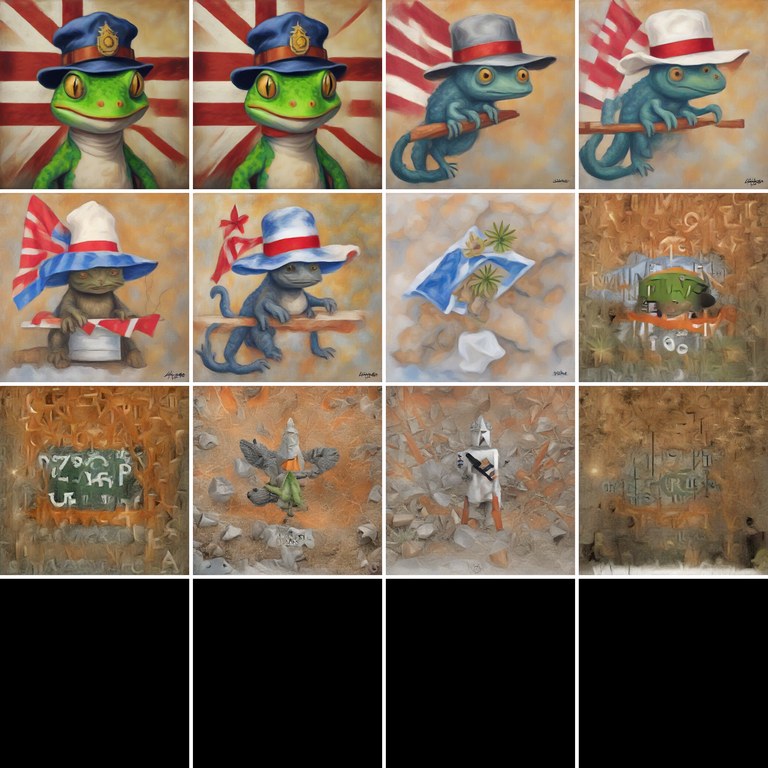}
            {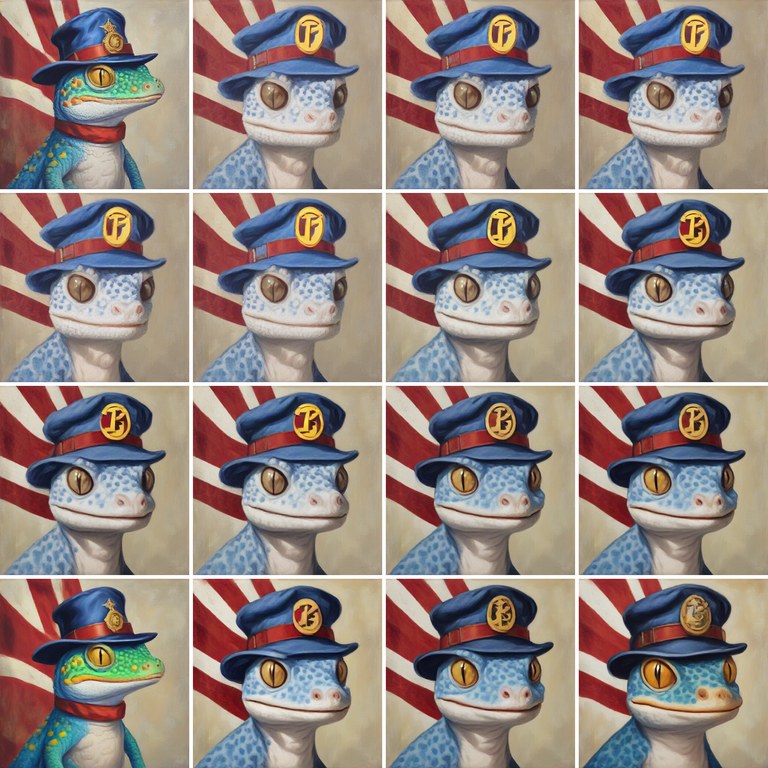}
            {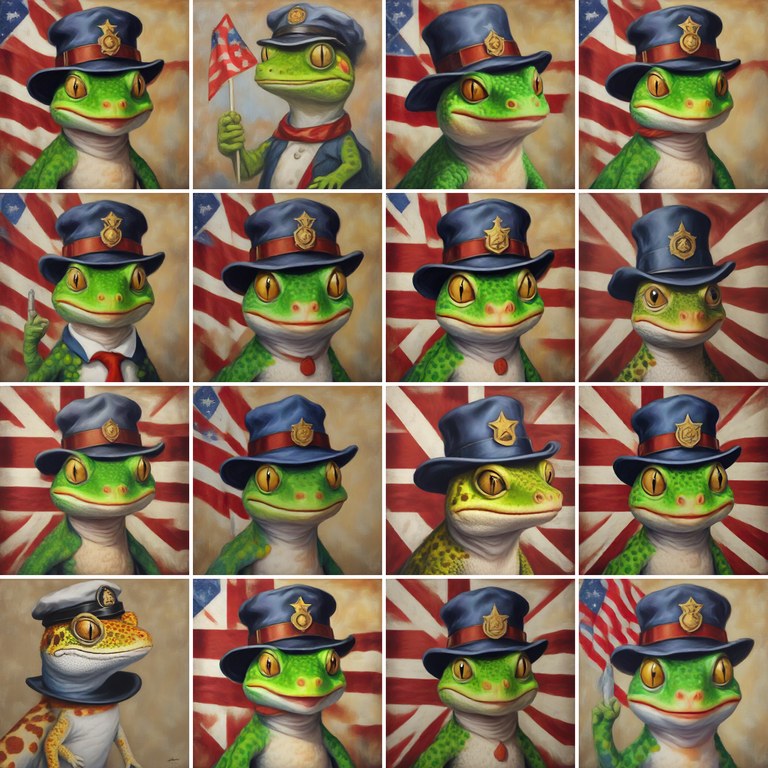}
            {Portrait of a gecko wearing a train conductor’s hat and holding a flag that has a yin-yang symbol on it. Oil on canvas.}
        
        \end{minipage}
    }
    \vspace{-10pt} %
    \caption{SDXL-Turbo Top-16 images with highest fitness for sample prompts.}
    \label{fig:app_framework_images}
\end{figure}

\begin{figure}[htbp]
    \centering
    \scalebox{0.9}{
        \begin{minipage}{\textwidth}
        
        \comparisonBlockSmall{baseline}
            {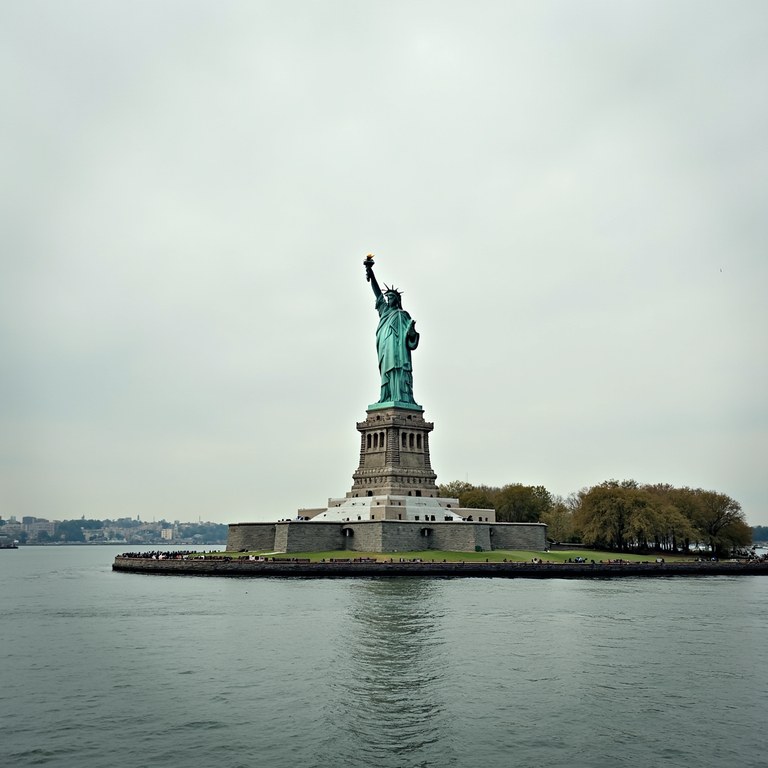}
            {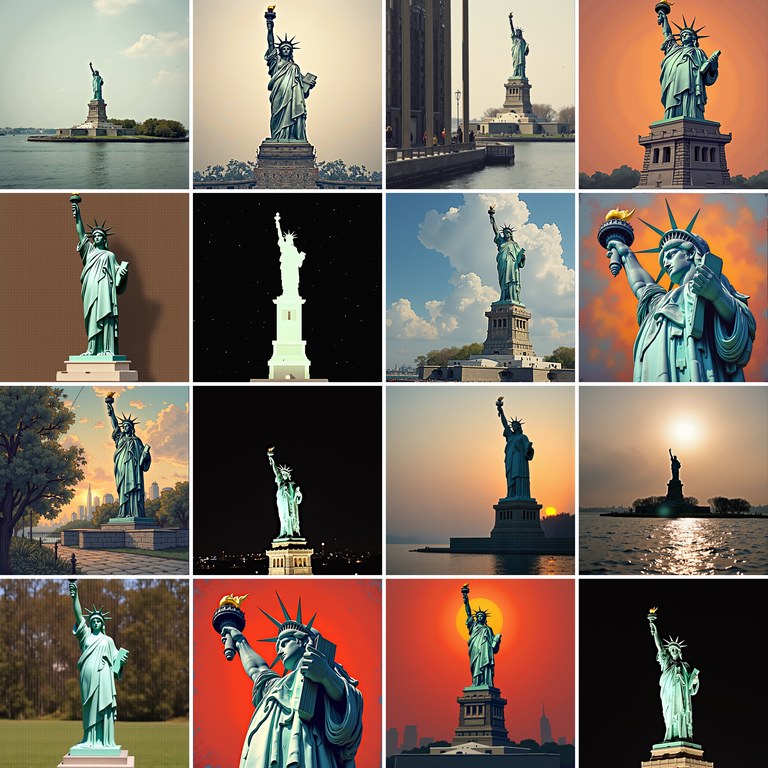}
            {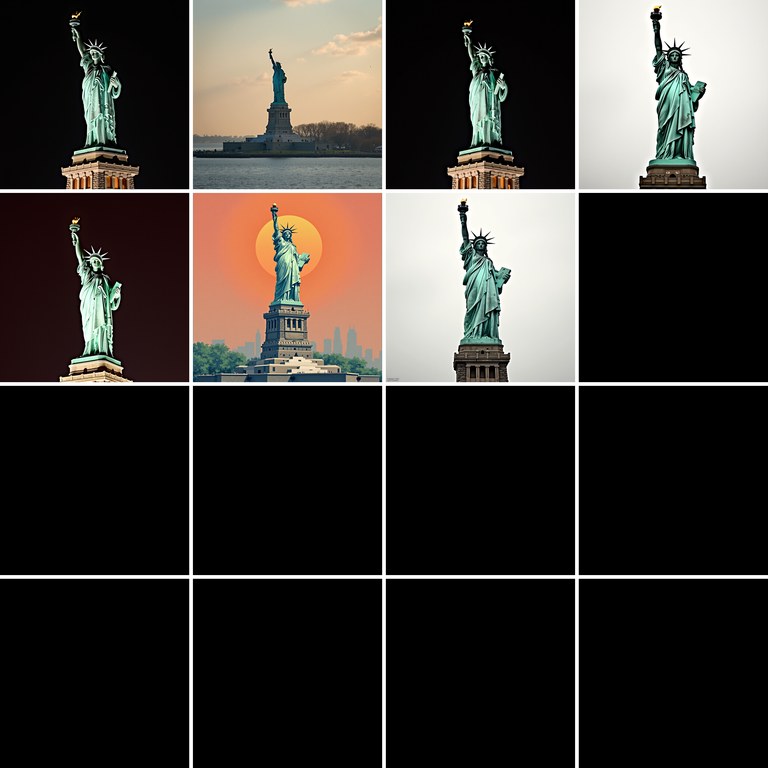}
            {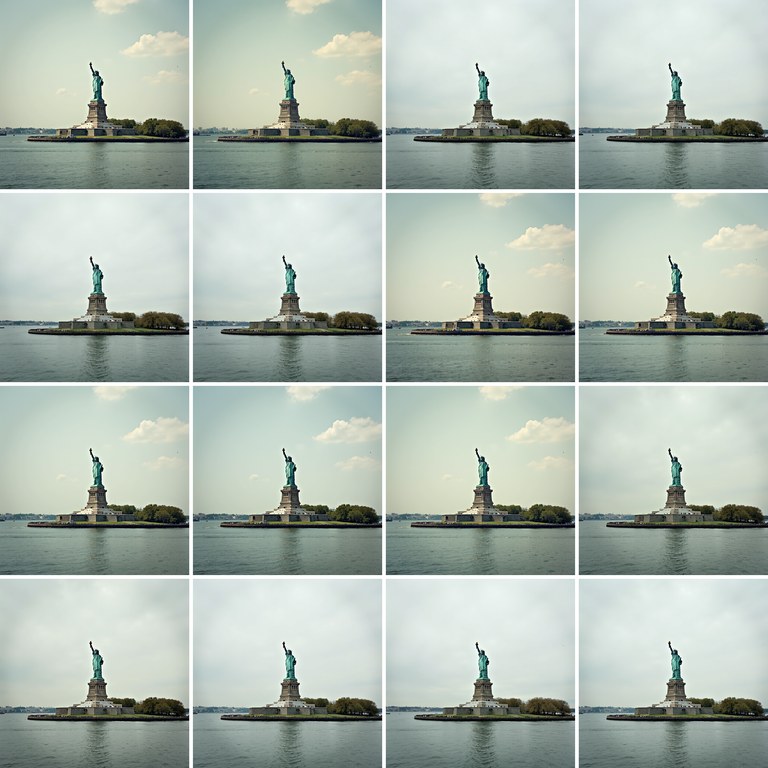}
            {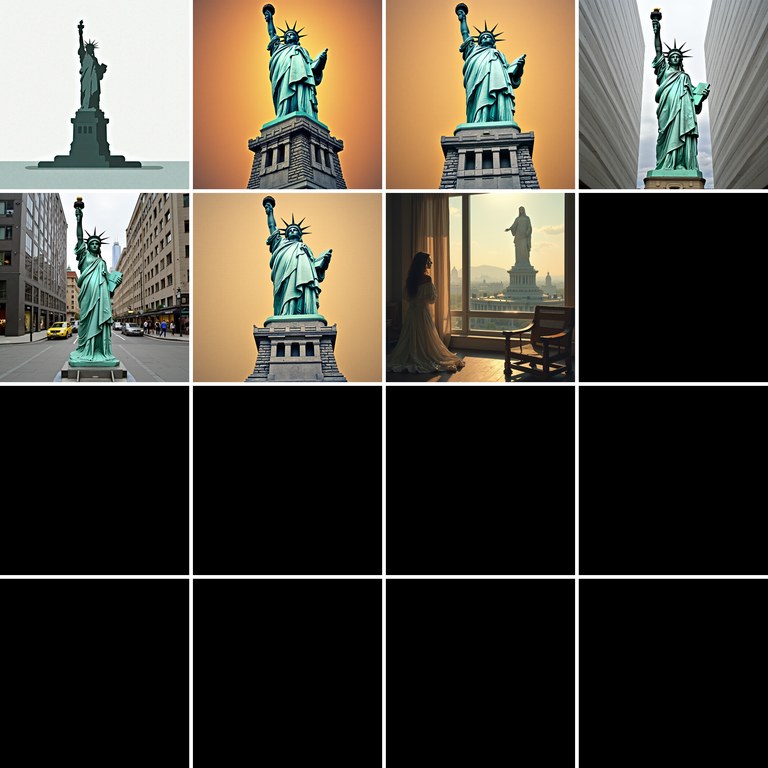}
            {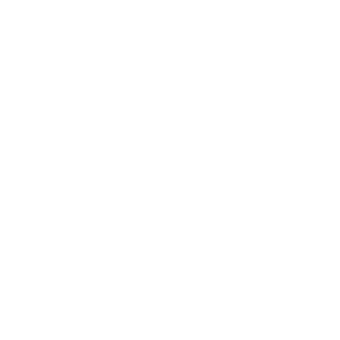} 
            {figures/flux_images/white}
            {The Statue of Liberty}

        \comparisonBlockSmall{baseline}
            {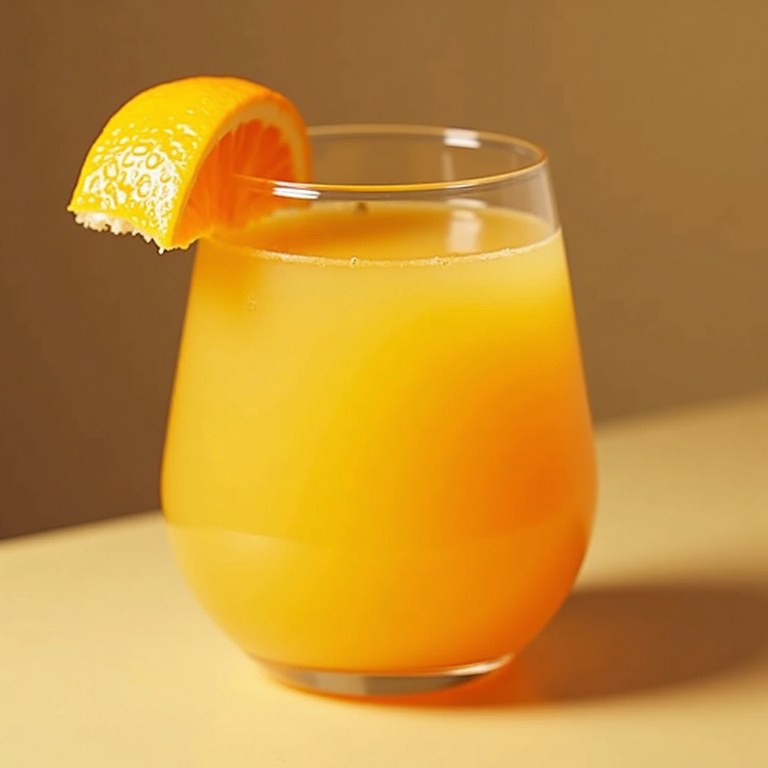}
            {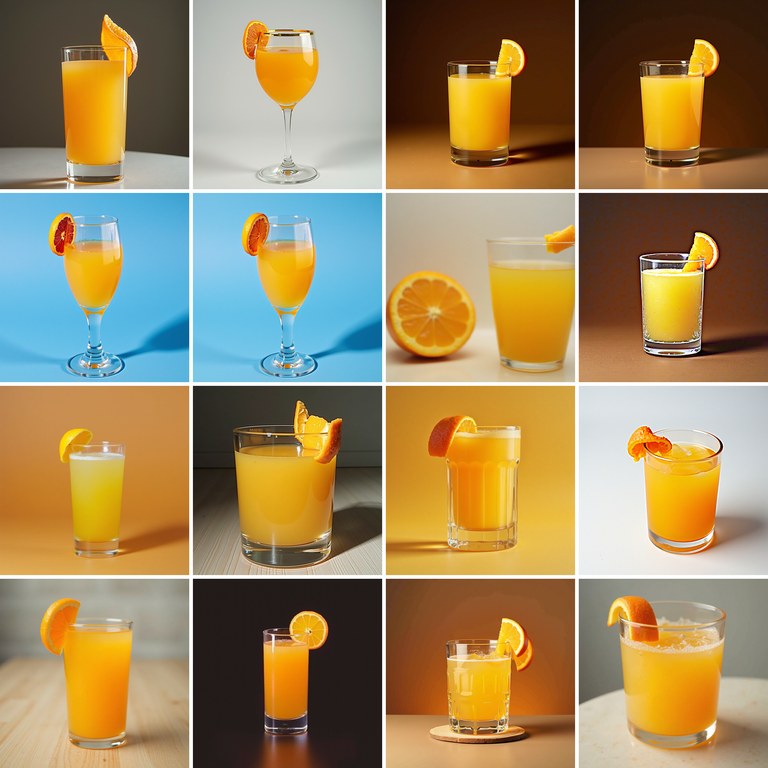}
            {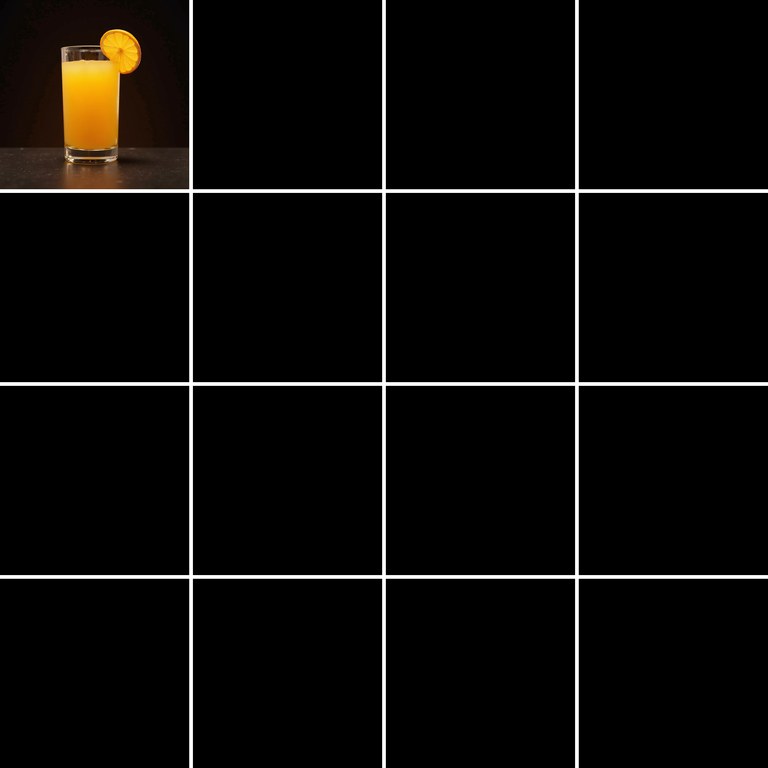}
            {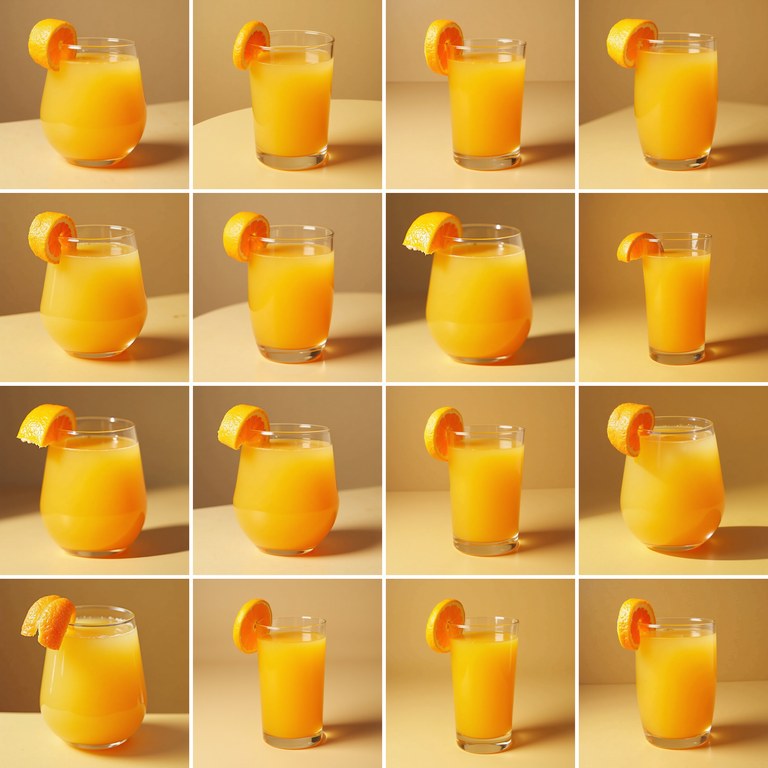}
            {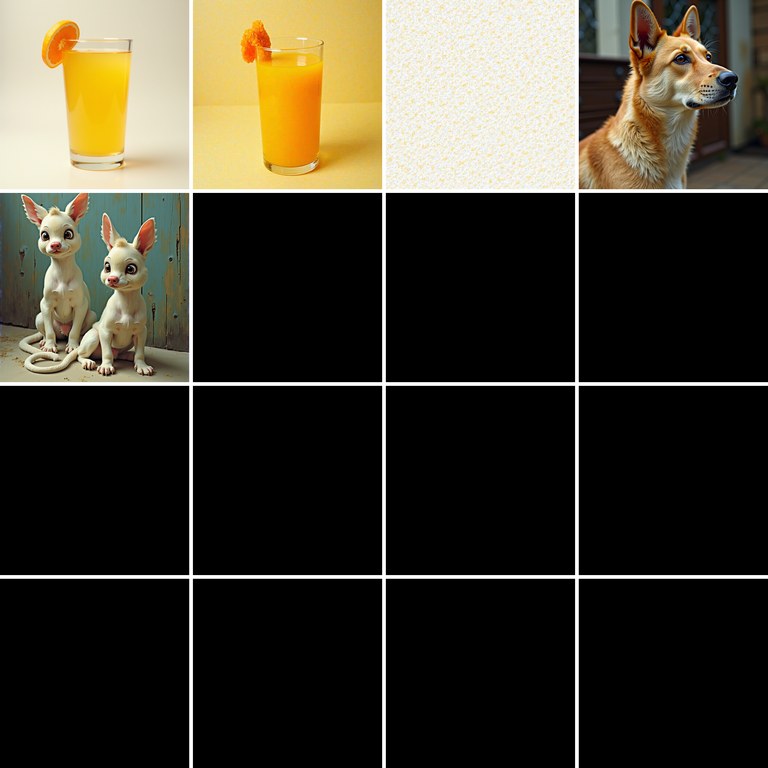}
            {figures/flux_images/white}
            {figures/flux_images/white}
            {a glass of orange juice with an orange peel stuck on the rim}
        
        \end{minipage}
    }
    \vspace{-10pt}
    \caption{Flux Top-16 images with highest fitness for sample prompts.}
    \label{fig:app_flux_framework_images}
\end{figure}

\Cref{fig:app_framework_images} and \Cref{fig:app_flux_framework_images} shows the images generated by our framework baselines and the corresponding seed image for sample prompts, for SDXL-Turbo and Flux respectively.

\subsection{Variation Operator Experiment Images}\label{app:images_variation}

\begin{figure}[htbp]
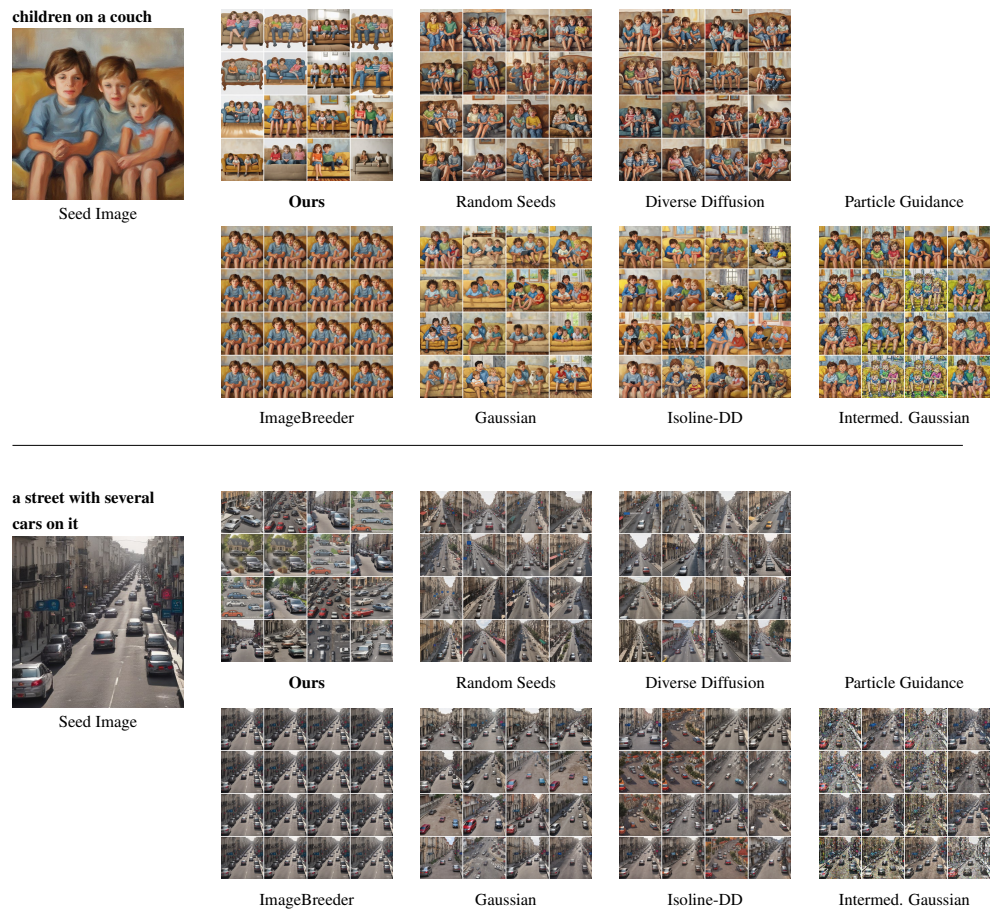

    \centering
    \scalebox{0.9}{
        \begin{minipage}{\textwidth}

        \comparisonBlock{variation}
            {figures/appendix_variation_images/18-initial}
            {figures/appendix_variation_images/18-dns_svd}
            {figures/appendix_variation_images/18-dns_random}
            {figures/appendix_variation_images/18-dns-diverse-diffusion}
            {figures/flux_images/white}
            {figures/appendix_variation_images/18-dns_mixed}
            {figures/appendix_variation_images/18-dns_gaussian}
            {figures/appendix_variation_images/18-dns_isoline}
            {figures/appendix_variation_images/18-dns_hidden_gaussian}
            {children on a couch}

        \comparisonBlock{variation}
            {figures/appendix_variation_images/31-initial}
            {figures/appendix_variation_images/31-dns_svd}
            {figures/appendix_variation_images/31-dns_random}
            {figures/appendix_variation_images/31-dns-diverse-diffusion}
            {figures/flux_images/white}
            {figures/appendix_variation_images/31-dns_mixed}
            {figures/appendix_variation_images/31-dns_gaussian}
            {figures/appendix_variation_images/31-dns_isoline}
            {figures/appendix_variation_images/31-dns_hidden_gaussian}
            {a street with several cars on it}

        \end{minipage}
    }
    \vspace{-10pt}
    \caption{SDXL-Turbo Top-16 images with highest fitness for sample prompts for Variation Operator Experiments.}
    \label{fig:app_variation_images}
\end{figure}

\begin{figure}[htbp]
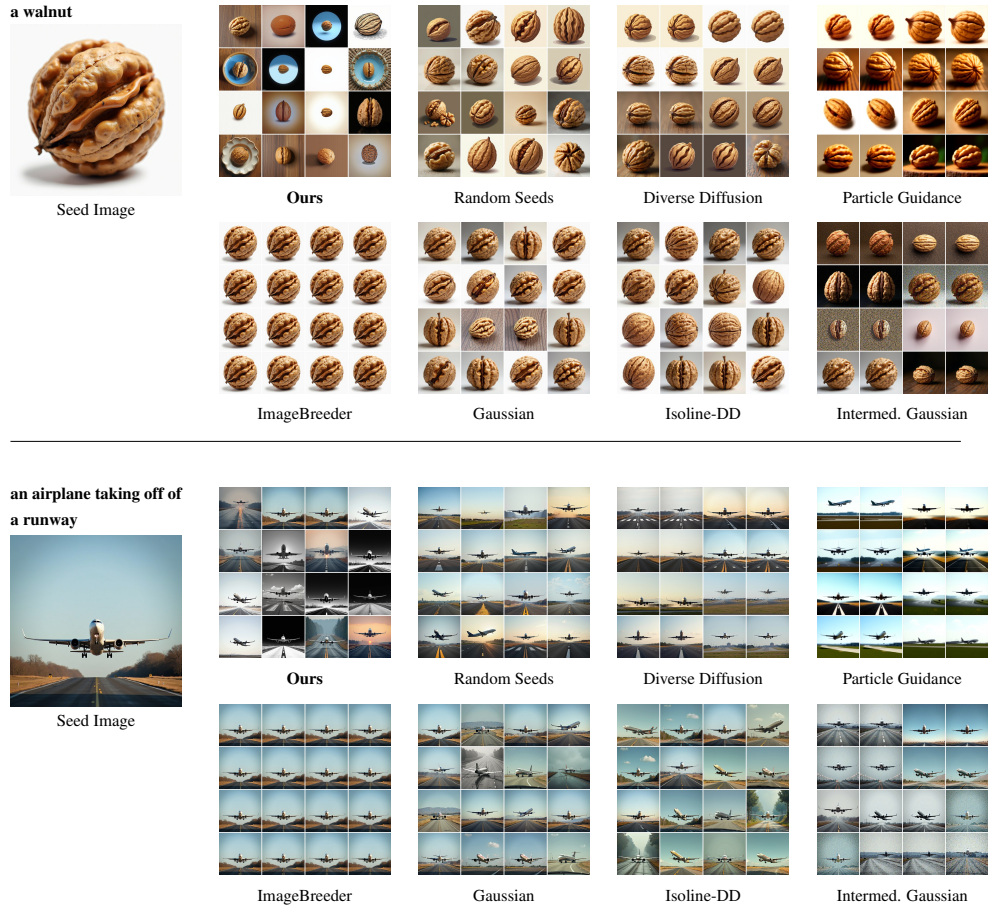

    \centering
    \scalebox{0.9}{
        \begin{minipage}{\textwidth}

        \comparisonBlock{variation}
            {figures/flux_images/28-initial}
            {figures/flux_images/28-dns}
            {figures/flux_images/28-random}
            {figures/flux_images/28-diverse-diffusion}
            {figures/flux_images/28-particle-guidance}
            {figures/flux_images/28-imagebreeder}
            {figures/flux_images/28-gaussian}
            {figures/flux_images/28-isoline}
            {figures/flux_images/28-hidden-gaussian}
            {a walnut}

        \comparisonBlock{variation}
            {figures/flux_images/4-initial}
            {figures/flux_images/4-dns}
            {figures/flux_images/4-random}
            {figures/flux_images/4-diverse-diffusion}
            {figures/flux_images/4-particle-guidance}
            {figures/flux_images/4-imagebreeder}
            {figures/flux_images/4-gaussian}
            {figures/flux_images/4-isoline}
            {figures/flux_images/4-hidden-gaussian}
            {an airplane taking off of a runway}

        \end{minipage}
    }
    \vspace{-10pt}
    \caption{Flux top-16 images with highest fitness for sample prompts for Variation Operator Experiments.}
    \label{fig:app_flux_variation_images}
\end{figure}

In \Cref{fig:app_variation_images} and \Cref{fig:app_flux_variation_images} we include the results images generated by our variation operator baselines and the seed image for sample prompts, for SDXL-Turbo and Flux respectively.

\clearpage
\newpage
\section{Full List of Prompts}\label{app:prompts_list}

\begin{table}[ht!]
\centering
\caption{List of prompts with categories and challenge levels.}
\resizebox{\textwidth}{!}{%
\begin{tabular}{@{} c l l p{11cm} @{}}
\toprule
\textbf{ID} & \textbf{Category} & \textbf{Challenge Level} & \textbf{Prompt} \\
\midrule
1 & Abstract & Basic & happiness \\
2 & Abstract & Basic & metal \\
3 & Abstract & Basic & element \\
4 & Vehicles & Simple Detail & an airplane taking off of a runway \\
5 & Vehicles & Fine-grained Detail & an antique car by a beach \\
6 & Vehicles & Fine-grained Detail & a drop-top sports car coming around a bend in the road \\
7 & Illustrations & Imagination & a cube made of porcupine \\
8 & Illustrations & Basic & a musical note \\
9 & Illustrations & Style \& Format & a sketch of a skyscraper \\
10 & Arts & Complex & an oil surrealist painting of a dreamworld on a seashore where clocks and watches appear to be inexplicably limp and melting in the desolate landscape. a table on the left, with a golden watch swarmed by ants. a strange fleshy creature in the center of the painting \\
11 & Arts & Complex & A raccoon wearing formal clothes, wearing a tophat and holding a cane. The raccoon is holding a garbage bag. Oil painting in the style of Vincent Van Gogh. \\
12 & Arts & Style \& Format & an abstract painting of a tree and a building \\
13 & World Knowledge & Basic & The Statue of Liberty \\
14 & World Knowledge & Basic & the grand canyon \\
15 & World Knowledge & Complex & A portrait of a metal statue of a pharaoh wearing steampunk glasses and a leather jacket over a white t-shirt that has a drawing of a space shuttle on it. \\
16 & People & Fine-grained Detail & a knight holding a long sword \\
17 & People & Fine-grained Detail & an elder politician giving a campaign speech \\
18 & People & Simple Detail & children on a couch \\
19 & Animals & Writing \& Symbols & Portrait of a tiger wearing a train conductor’s hat and holding a skateboard that has a yin-yang symbol on it. woodcut \\
20 & Animals & Basic & a cat \\
21 & Animals & Style \& Format & Portrait of a gecko wearing a train conductor’s hat and holding a flag that has a yin-yang symbol on it. Oil on canvas. \\
22 & Artifacts & Fine-grained Detail & a yellow diamond-shaped sign with a deer silhouette \\
23 & Artifacts & Writing \& Symbols & a black t-shirt with the peace sign on it \\
24 & Artifacts & Simple Detail & a yield sign \\
25 & Food \& Beverage & Fine-grained Detail & a glass of orange juice with an orange peel stuck on the rim \\
26 & Food \& Beverage & Simple Detail & a roast turkey on the table \\
27 & Food \& Beverage & Imagination & A castle made of tortilla chips, in a river made of salsa. There are tiny burritos walking around the castle \\
28 & Produce \& Plants & Basic & a walnut \\
29 & Produce \& Plants & Style \& Format & A photo of a palm tree made of water. \\
30 & Produce \& Plants & Complex & a tree reflected in the hood of a blue car \\
31 & Outdoor Scenes & Simple Detail & a street with several cars on it \\
32 & Outdoor Scenes & Quantity & two chemtrails forming an X in blue sky \\
33 & Outdoor Scenes & Linguistic Structures & a house with no windows \\
34 & Indoor Scenes & Complex & An empty fireplace with a television above it. The TV shows a lion hugging a giraffe. \\
35 & Indoor Scenes & Fine-grained Detail & a wood cabin with a fire pit in front of it \\
36 & Indoor Scenes & Quantity & a bunch of laptops piled on a sofa \\
\bottomrule
\end{tabular}%
}
\end{table}

\section{Wall-clock Time Experiments}\label{app:wallclock_experiments}

In this section, we provide a detailed analysis of the computational cost of the $h$-space Spectral Mutation operator.
This analysis was performed on 12 seeds, with one seed selected from each prompt category.
We exclude CVT-ME and QDHF from this comparison as both utilise our Spectral Mutation operator as their primary variation mechanism.

\subsection{Per-Image Generation Time}

The median time per image generation for each method is provided in \Cref{tab:operator_time}.While most evolutionary operators rely on simple additive noise, our approach involves a Jacobian-based spectral estimation which adds meaningful overhead relative to the base model's forward pass.

\begin{table}[htbp]
\centering
\scriptsize
\setlength{\tabcolsep}{6pt}
\renewcommand{\arraystretch}{1.2}
\caption{Computational Overhead: Median Time per Image (Seconds)}
\label{tab:operator_time}
\begin{tabular}{|l|c|c|}
\hline
Algorithm & SDXL-Turbo (s) & Flux (s) \\ \hline
Spectral-Guided (Ours) & 0.78 & 14564.48 \\ \hline
Sep-CMA-ES & 0.002 & 7.22 \\ \hline
CMA-MAE & 0.001 & 6.96 \\ \hline
Adam & 0.90 & - \\ \hline
Random Seeds & 0.0003 & 1.37 \\ \hline
Diverse Diffusion & 0.002 & 16.39 \\ \hline
Particle Guidance & - & 18885.01 \\ \hline
ImageBreeder & 0.002 & 7.85 \\ \hline
Gaussian & 0.002 & 6.36 \\ \hline
Isoline-DD & 0.002 & 6.84 \\ \hline
Intermed.-layer Gaussian & 0.003 & 5864.6 \\ \hline

\end{tabular}
\end{table}

On SDXL-Turbo, our Spectral operator is the most expensive mutation other than Adam, which requires a full backward pass through the model.
On Flux.1, our per-image cost ($14{,}564$s) is of the same order as the guidance-based diversity method, Particle Guidance ($18{,}885$s),
and is roughly $2.5\times$ that of the Intermediate-layer Gaussian operator ($5{,}865$s).
It remains several orders of magnitude greater than the simple additive-noise operators, such as Random Seeds, Gaussian, and Isoline-DD, whose mutations cost only a few seconds per image.

We acknowledge that this overhead is a key limitation of our work: the Jacobian-based spectral estimation is substantially more costly than unstructured noise-based mutation.
We note, however, that this expense is characteristic of methods that actively steer the generative trajectory rather than perturbing latents with unstructured noise.
The baseline whose cost is comparable to ours, Particle Guidance, are precisely those designed to promote diversity through guidance, which suggests that the overhead largely reflects the general cost of structured, semantically guided exploration rather than an inefficiency specific to our operator.
Whether this cost is warranted ultimately depends on the quality and diversity of the resulting solutions, which we examine under a fixed evaluation budget in \Cref{sec:framework_results} and \Cref{sec:variation_results}.
Reducing this overhead, for example through cheaper approximations to the spectral
decomposition, is a promising direction for future work.

Finally, we note that the significance of this overhead is context-dependent.
The per-image cost is most salient in settings where image generation is the principal bottleneck and candidates are cheap to evaluate.
In many scientific design applications, however, the cost of generation is small relative to the cost of downstream validation: in domains such as biological or materials design, for instance, evaluating a candidate may involve wet-lab synthesis, assays, or physical testing that can dominate the overall time and resource budget.
In such regimes, the additional generation cost of our operator is likely to be a minor contribution to the total, and the benefit of surfacing a more diverse set of candidates, thereby reducing the number of expensive validation cycles needed to find a viable solution,
may well outweigh it.
We do not evaluate such applications here, and the trade-off will depend on the specific costs involved, but we highlight this as a setting in which the computational profile of our method may be more favourable.

\section{Impact Statement}\label{app:impact}

This paper presents work whose goal is to advance the field of machine learning, specifically at the intersection of Evolutionary Computation and Generative AI. Our proposed framework, Spectral-Guided Quality-Diversity, enables the systematic exploration of the latent spaces of large-scale generative models. This capability has significant potential for positive societal impact, particularly in scientific discovery, where it can accelerate the design of novel proteins, materials, or drugs by generating diverse portfolios of candidates that might otherwise be missed by mode-seeking methods. Furthermore, by explicitly optimising for diversity, our approach contributes to bias mitigation, helping to uncover regions of the data distribution that are often suppressed by standard sampling, potentially surfacing underrepresented groups or concepts and counteracting mode-collapse-induced bias.

However, we acknowledge that methods designed to uncover ``hidden" diversity inherently carry risks. By navigating away from the high-likelihood modes of the training distribution, such algorithms could potentially circumvent safety alignment measures designed to suppress harmful, offensive, or copyrighted content. We encourage researchers and practitioners employing this framework to integrate robust post-generation safety filtering, particularly when deploying these methods in open-ended or user-facing applications.

\end{document}